\documentclass[journal]{IEEEtran}
\ifCLASSINFOpdf
\else
\fi

\usepackage{times}
\usepackage{epsfig}
\usepackage{graphicx}
\usepackage{amsmath}
\usepackage{amssymb}
\usepackage{algorithm}
\usepackage{caption}
\usepackage{algorithmic}
\usepackage{float}
\usepackage{soul}
\usepackage{xspace}
\usepackage{xcolor}
\usepackage{wrapfig}

\usepackage{mathtools}
 \usepackage{multirow}

\usepackage[justification=centering]{caption}

\def\onedot{\ifx\@let@token.\else.\null\fi\xspace}

\def\eg{\emph{e.g}\onedot}

\newcommand{\old}[1]{}
\newcommand{\new}[1]{\textcolor[rgb]{0,0,0}{#1}}

\newcommand{\anran}[1]{\textcolor[rgb]{0,0,0}{#1}}

\begin{document}
%
\title{One Sketch for All: One-Shot Personalized Sketch Segmentation}
%
%
%

\author{Anran~Qi$^1$,  Yulia~Gryaditskaya$^{1,2}$, Tao~Xiang$^1$, Yi-Zhe~Song$^1$\\
SketchX Lab, CVSSP$^1$,  Surrey Institute for People-Centred AI$^2$\\University of Surrey \\
}
\maketitle

\begin{abstract}
We present the first one-shot personalized sketch segmentation method. 
We aim to segment all sketches belonging to the same category provisioned with a single sketch with a given part annotation while (i) preserving the parts semantics embedded in the exemplar, and (ii) being robust to input style and abstraction. 
We refer to this scenario as \emph{personalized}. 
With that, we importantly enable a much-desired personalization capability for downstream fine-grained sketch analysis tasks. 
To train a robust segmentation module, we deform the exemplar sketch to each of the available sketches of the same category. 
Our method generalizes to sketches not observed during training. 
Our central contribution is a sketch-specific hierarchical deformation network. Given a multi-level sketch-strokes encoding obtained via a graph convolutional network, our method estimates rigid-body transformation from the target to the exemplar, on the upper level. Finer deformation from the exemplar to the globally warped target sketch is further obtained through stroke-wise deformations, on the lower-level. Both levels of deformation are guided by mean squared distances between the keypoints learned without supervision, ensuring that the stroke semantics are preserved.
We evaluate our method against the state-of-the-art segmentation and perceptual grouping baselines re-purposed for the one-shot setting and against two few-shot 3D shape segmentation methods.
We show that our method outperforms all the alternatives by more than $10\%$ on average. 
Ablation studies further demonstrate that our method is robust to \emph{personalization}: changes in input part semantics and style differences.
\end{abstract}

{\let\thefootnote\relax\footnotetext{This work has been submitted to the IEEE for possible publication. Copyright may be transferred without notice, after which this version may no longer be accessible.}}

%
\IEEEpeerreviewmaketitle


\section{Introduction}
With the appearance of large-scale sketch datasets and recent advances in deep learning, sketch-related research thrives \cite{ Liu_2019_ICCV, 
Ghosh_2019_ICCV, Gao_2020_CVPR,Zheng_2020_CVPR,Lin_2020_CVPR, Liu_2020_ECCV, Liu_F_2020_ECCV}. 
Sketch segmentation in particular is an important capability that underpins the recent focus of fine-grained sketch analysis, such as part-based sketch-based modeling and retrieval \cite{jones2020shapeassembly, li2016fine} \old{or}\anran{and} fine-grained sketch editing \cite{Zou_2018_ECCV, su2020sketchhealer}. 

Existing methods aiming at semantic sketch segmentation rely on the availability of large-scale carefully annotated sketch datasets. 
However, obtaining such annotations for new categories is an extremely labor-intensive task. 
\emph{\old{Moreover, current datasets provide fixed part sketches labeling, disregarding the subjective nature of segmentations}
\new{Moreover, current datasets contain a single set of labels per sketch, disregarding the subjective nature of the segmentation task} \cite{MartinFTM01,perteneder2015cluster} -- 
this impedes downstream applications, not allowing \old{the number of meaningful labels}\new{the labels} to be task-specific.} 

\begin{figure}[t]
\vspace{-6pt}
\begin{center}
   \includegraphics[width=1.0\linewidth]{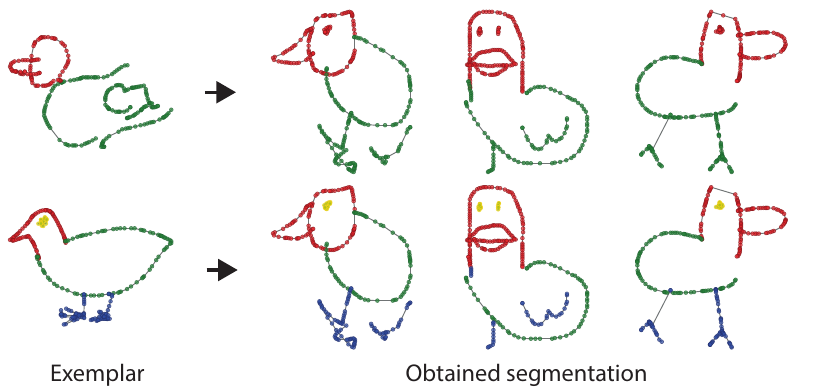}
\end{center}
\vspace{-6pt}
   \caption{Given an exemplar annotated sketch, we are able to transfer \old{the}\new{its} labels to any \old{amount}\new{number} of sketches of the same category and achieve task-specific segmentation.
}
\label{fig:teaser}
\vspace{-8pt}
\end{figure}

In this paper, for the first time, we address the problem of \emph{personalized} sketch segmentation under a one-shot setting, as an attempt to facilitate the fine-grained analysis of sketches. 
Namely, given \old{\emph{one} input sketch that is part annotated}\new{\emph{one} annotated sketch}, we seek to label any number of gallery sketches in a consistent manner \old{by conforming}\new{according} to the same user-defined semantic interpretation, and robustly to differences in drawing style and abstraction between the input and gallery.
We refer to this scenario as \emph{personalized sketch segmentation}.
Fig.~\ref{fig:teaser} offers two working examples where two sketches of different styles and part decomposition were used to segment unseen sketches of the same category consistent with each input semantic segmentation.

The problem of one-shot personalized sketch segmentation is however non-trivial. 
The key challenge lies with deriving means \old{of effectively transferring}\new{to efficiently transfer} the semantic labels in the exemplar to all target sketches so that parts semantics are preserved. 
Diversity in sketch depictions among humans (e.g., the ducks in Fig.~\ref{fig:teaser}) makes this task even more challenging.

In our work, we exploit \emph{deformation}, and train the segmentation module by morphing the exemplar sketch to available sketches of a given category. 
To increase robustness, we propose a hierarchical deformation model specific to sketch data. During training, we first globally align all sketches with the exemplar\new{,} and then \old{obtain}\new{predict} a fine-level warping of the exemplar towards \old{the}\new{each of} globally aligned sketches.
Such an approach simplifies the task of the segmentation network: The segmentation network does not need to deal with large geometric differences.
\old{At inference time}\new{During inference}, we first perform a global alignment with the exemplar, then \new{we} predict the segmentation and\new{,} finally \new{we} apply an inverse global transformation operator.

Our first key contribution is in exploiting the connectivity information encoded in sketch strokes, in both parts of our segmentation model: deformation and segmentation. To embed this information we obtain a multi-level sketch encoding using a graph convolution network (GCN). The graph consists of dynamic and static edges. The role of dynamic edges is to encode global information embedded into the holistic sketch. The static edges connect the consequent points along strokes, encoding stroke-level information. \old{Such representation allows to retain both sketch- and stroke-level embeddings in a synergistic fashion.}

The sketch deformation module, \old{being}\new{which is} our central contribution, builds around this sketch representation. On the holistic sketch level, we first predict a global rigid-body transformation with respect to an exemplar sketch, relying on an analytic solver. 
We observe that the conventional Chamfer distance \cite{wang2020few,yuan2020ross} between an input sketch and an exemplar is not sufficient to predict the global transformation which preserves semantic part correspondences, due to the sparse nature of sketches. 
We instead jointly train for unsupervised keypoints prediction \cite{chen2020unsupervised} which are used to derive global transformation. 
Such rigid-body deformation however only \textit{roughly} aligns the gallery sketch to the exemplar. 
To achieve finer alignment, \old{we impose stroke-level deformations rather than individually on stroke points}\new{we apply deformations at the stroke level, and not individually on the stroke points}. 
To avoid unrealistic stroke distortions\new{,} we limit the transformations to \old{stroke-level} rotation, translation\new{,} and scaling \new{at the stroke level}. 
To achieve a globally consistent deformation \old{on}\new{at} this level, we leverage both Chamfer distance and mean square error \new{computed} on keypoints \new{positions}. 
The distances are computed between the \new{keypoints of the} fine-level deformed exemplar and a holistic sketch-level deformed gallery sketch.

Given the deformed exemplar\new{,} we train a segmentation module in a standard fashion via a cross-entropy loss, which given the full sketch encoding predicts a label for each sketch point. We \old{further}\new{also} observe that sketch strokes are frequently \old{fully}\new{completely} contained within one semantic part. 
\old{We thus}\new{Therefore, we} further condition the segmentation label predictor on the stroke encoding, eliminating the need for a labels refinement step such as graph-cuts \cite{li2018fast} or conditional random fields \cite{schneider2016example, zhu20202d}.

In summary, (i) we propose for the first time the problem of one-shot sketch segmentation; (ii) we show that by transferring the semantic labeling from an input sketch to target ones, a much-desired personalized segmentation capability can be achieved; (iii) we propose a hierarchical sketch deformation framework that faithfully deforms the input sketch to each reference, as means to assist the transfer; (iv)  we conduct multiple ablation studies demonstrating the robustness of our proposed method in terms of variations in part semantics and sketching style; (v) we show an advantage of our method over alternative solutions for a few-shot sketch segmentation.

\section{Related work}

\label{sec:related_work}
\paragraph{Sketch segmentation}
Prior works on sketch segmentation can be divided into two categories based on the segmentation goals: strokes perceptual grouping \cite{qi2013sketching, qi2015making, li2018universal, liu2018strokeaggregator} and semantic segmentation 
\cite{sun2012free, huang2014data, schneider2016example, zhu2018part, li2018fast, wu2018sketchsegnet, wang2018sketchpointnet,  li2019toward, qi2019sketchsegnet+, wang2019spfusionnet, hahnlein2019bitmap, yang2020sketchgcn, zhu2020sketchppnet}.
Our work belongs to the second category. \old{Yet, we}\new{However}, for the first time, \new{we} consider the problem of \old{a }\emph{one-shot} semantic sketch segmentation.

\old{A}\new{The a}ppearance of large-scale annotated sketch datasets \cite{huang2014data, li2018universal, eitz2012humans, li2018fast} fostered research on supervised semantic sketch segmentation \old{with}\new{using} deep learning.
These methods can be classified into several groups \old{by the used sketch representation}\new{according to the sketch representation used}: 
image-based  \cite{li2018fast, zhu20202d, ge2020creative},
point-based  \cite{ hahnlein2019bitmap, wang2020multi},
ordered point sequences-based \cite{wu2018sketchsegnet, qi2019sketchsegnet+}, 
joint pixel-/point-based \cite{wang2019spfusionnet} methods,
and graph-based \cite{yang2020sketchgcn}.
\emph{Image-based} methods represent a sketch as a raster image\old{,} and build on \old{a}\new{the} success of convolutional neural networks (CNNs) in learning descriptive features. 
Zhu et al.~\cite{zhu20202d} combine\anran{d} a CNN-based segmentation with \old{a }CRF-based \old{refining}\new{refinement}.
\emph{Point-based} methods \cite{ hahnlein2019bitmap, wang2020multi} represent\anran{ed} a sketch as an ordered point cloud. 
Such works \old{build}\anran{built} on the point-cloud architectures, first designed for \new{the} 3D shape analysis and representation \cite{qi2017pointnet, qi2017pointnet++, li2018pointcnn, wang2019dynamic}. 
Point-based representation \old{allows to reduce models complexity,}\new{reduces the complexity of models} compared to image\old{d}-based representations\old{,} due to the sparsity of lines in sketches. 
For the task of sketch recognition, Wang et al.~\cite{wang2018sketchpointnet} proposed a sketch-dedicated point cloud architecture. 
Both Wang et al.~\cite{wang2018sketchpointnet} and Hahnlein et al.~\cite{ hahnlein2019bitmap} \old{take}\anran{took} as points features not only \new{the} points spatial coordinates\old{,} but also \new{the} strokes order.
A number of works \cite{wu2018sketchsegnet, qi2019sketchsegnet+,yang2020s} exploit\anran{ed} \emph{recurrent neural networks (RNNs)} to translate sequence of strokes into their semantic parts. 
Such architectures were first designed for \old{the }the task of sketch recognition \cite{ha2017neural, li2018sketch, jia2020coupling}. 
\old{We though}\new{However, we} did not observe a correlation between a semantic stroke label and a stroke number in \old{a}\new{the} general case.
Wang et al.~\cite{wang2019spfusionnet} \emph{\old{fuses}} \emph{\anran{fused}} a prediction of \old{the}\new{a} dedicated image-based architecture with the one obtained from a point-based architecture \cite{qi2017pointnet}. 
Recently, Zhu et al.~\cite{zhu2020sketchppnet} applied a similar idea to \old{a}\new{the} sketch recognition task.
\old{Despite \emph{graph-based} sketch representation was commonly used in earlier works on sketch analysis, just recently graph convolution networks (GCNs) were adopted for sketch processing.}
\new{While earlier sketch analysis work commonly used \emph{graph-based} representation of sketches, it is only recently that Graph Convolution Networks (GCNs) have been adopted for sketch processing.}
Yang et al.~\cite{yang2020sketchgcn} proposed a two branches GCN for a supervised sketch segmentation task, implementing graph convolutions as was proposed in \cite{wang2019dynamic} and \cite{li2019deepgcns}.
Su et al.~\cite{su2020sketchhealer} exploited GCN with a sparse number of nodes for the partial sketch completion task.
Yang et al.~\cite{yang2020s} leverage\anran{d} \new{a} RNN \old{with}\new{in} conjunction with \new{a} GCN for \old{a }sketch recognition\old{ task}.
In our work, we \old{leverage}\new{employ} a GCN for \old{the task of a }few-shot sketch segmentation, leveraging a multilevel sketch encoding.


\paragraph{3D shape segmentation}
Concurrently, two approaches for \old{a }few-shot 3D shape segmentation \old{were}\new{have} recently \new{been} proposed \cite{yuan2020ross, wang2020few}, relying on the idea of being able to morph a template shape to an arbitrary target shape. 
Yuan et al.~\cite{yuan2020ross} directly transfer\old{s}\anran{ed} a label from the morphed template to the target shape by proximity. 
Wang et al.~\cite{wang2020few} instead learn\old{s}\anran{ed} a continuous probability distribution function that learns to assign to each point in space \old{a}\new{the} probability of having a certain semantic label\old{,} conditioned on a global shape feature vector. 
Yuan et al.~\cite{yuan2020ross} exploit\anran{ed} mesh connectivity, while Wang et al.~\cite{wang2020few} \old{rely}\anran{relied} on a point-based shape representation.

Chen et al.~\cite{chen2019bae} proposed an autoencoder for unsupervised consistent segmentation of shapes from the same class\old{,} and demonstrated how such \new{an} architecture can be \old{tuned}\new{repurposed} for one-shot learning. 
The autoencoder consist\new{s} of several branches\old{, which} \new{that} are trained to encode complementary shape parts.
Instead of decomposing the shape into parts, Chen et al.~\cite{chen2020unsupervised} stud\old{y}\anran{ied} the problem of unsupervised prediction of semantically \old{consist}\new{consistent} keypoints across all shapes of the same class. 
Dense predicted keypoints correspondences are used to transfer semantic labels from one shape to another.

Our work extends these ideas to \old{a} few-shot sketch segmentation, taking into account point\old{s} connectivity and sketch sparsity, designed to be robust to arbitrary global sketch rotation and reflection. 

\begin{figure*}[t]
\begin{center}

\includegraphics[width=1.02\linewidth]{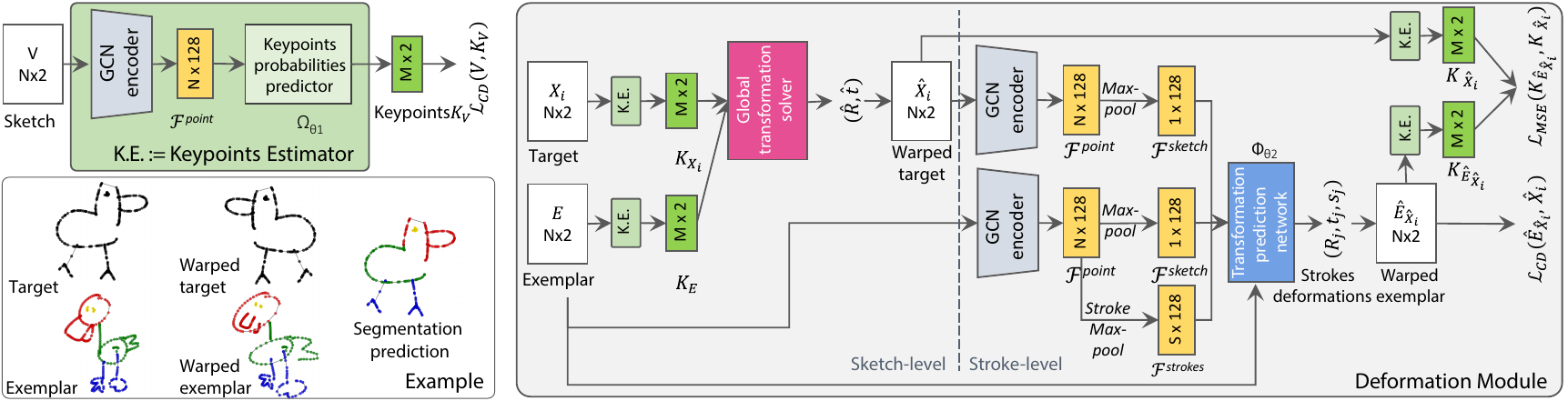}
\end{center}
\vspace{-10pt}
   \caption{Deformation module architecture (Section \ref{sec:deformation}) and a working example of the\old{ the} deformation process with \old{a}\new{the} segmentation prediction result by the segmentation module $\tau_{\theta_3}$, described in Section \ref{sec:segmentation}.}
\label{fig:training}
\vspace{-8pt}
\end{figure*}

\section{Method}

Our goal is, given an annotated exemplar sketch, to transfer its semantic part labels to an arbitrary target sketch of the same class as the exemplar sketch. 
We model this segmentation task as a two\new{-}step process, comprising exemplar morphing towards target sketches (Sec.~\ref{sec:deformation}) and a robust labels prediction (Sec.~\ref{sec:segmentation}). 
We leverage a graph convolutional network for multi-level sketch encoding (Sec.\ref{sec:sketch_representation}).

\subsection{Deformation model}
\label{sec:deformation}

We expect the deformation from \old{the}\new{an} exemplar to \old{the}\new{a} target to comply with the following rules: 
(i) The semantic meaning of \old{the}\new{a} stroke segment label should stay invariant under the deformation (\eg, a plane wing segment should not be deformed into a body\old{,} if they have different labels in the exemplar); 
(ii) The relative part relationship should \old{stay}\new{remain} invariant under the deformation (\eg, an eye should not move outside a head). 

We represent a sketch as an N-point set $V=\{v_i=(v^x_i, v^y_i)\}_{i=1,2,\ldots,N}$, where $v^x_i$ and $v^y_i$ are the 2D absolute coordinates of the point $v_i$. 

We model the deformation process  between the exemplar sketch $E$ and the unlabeled target sketch $X$ from the training set hierarchically. 
First, the global deformation\old{ that}\new{, which} accounts for \old{a}\new{the} global sketch rotation and reflection\new{,} aligns each sketch in the training batch with the exemplar. 
Then, the exemplar is morphed to each of the globally \old{morphed}\new{aligned} unlabeled sketches.
As mention\new{ed} in the introduction and shown in the ablation studies section, this hierarchical\old{ly} \old{two}\new{bi}-directional deformation allows to increase the accuracy of segmentation results \old{by simplifying the task of the segmentation module,} \new{by} lifting the requirement of learning rotation invariant segmentation.
The \old{full }deformation network is shown in Fig.~\ref{fig:training}\old{,} and explained in detail below. 

First, a sketch-level transformation, consisting of rotation/reflection $\hat{R} \subset  \mathbb{R}^{2 \times 2} $ and translation $\hat{\mathbf{t}} \subset \mathbb{R}^{2 \times 1} $, aligns an unlabeled sketch towards an exemplar sketch:
\begin{equation}
\hat{x}_i = \hat{R} x_i + \hat{\mathbf{t}},	\forall x_i \in X,   
\end{equation}
where the target sketch $X \subset \mathbb{R}^{2 \times {N}}$ consists of $N$ 2D stroke point coordinates. 
Then\old{,} $\hat{X} = \bigcup_{i} \hat{x}_i  \subset \mathbb{R}^{2 \times {N}}$ is the target sketch globally aligned with the exemplar sketch.

Second, a stroke-level transformation morphs the strokes of the exemplar sketch towards the globally aligned target sketch $\hat{X}$. 
The stroke\new{-}level deformation comprises per stroke rotation $R_{\hat{X}} \subset  \mathbb{R}^{2 \times 2} $, scaling $\mathbf{\sigma}_{\hat{X}}=[\sigma^x_{\hat{X}}, \sigma^y_{\hat{X}}] \subset  \mathbb{R}^{2} $ and translation $\mathbf{t}_{\hat{X}} \subset \mathbb{R}^{2 \times 1} $.
\old{The}\new{Thus, the} deformation model \old{thus can be}\new{is} written as follows:
\begin{equation}
\hat{e}_i = R_{j, \hat{X}}
\begin{bmatrix}
\sigma^x_{j, \hat{X}} & 0 \\
0  & \sigma^y_{j, \hat{X}}
\end{bmatrix}
e_i + \mathbf{t}_{j, \hat{X}},	\forall e_i \in \mathbf{s}_j \in E,   
\label{eq:deformed_template}
\end{equation}
where $\mathbf{s}_j$ is the $j$-th stroke of the the exemplar sketch $E\subset \mathbb{R}^{2 \times {N}}$, and $e_i$ is a\old{ 2D} stroke point coordinate. Then, $\hat{E}_{\hat{X}} = \bigcup_{i} \hat{e}_i  \subset \mathbb{R}^{2 \times {N}}$ is the exemplar sketch \old{aligned with}\new{morphed to} the globally aligned target sketch $\hat{X}$.

\begin{figure}[t]
\begin{center}
   \includegraphics[width=1.0\linewidth]{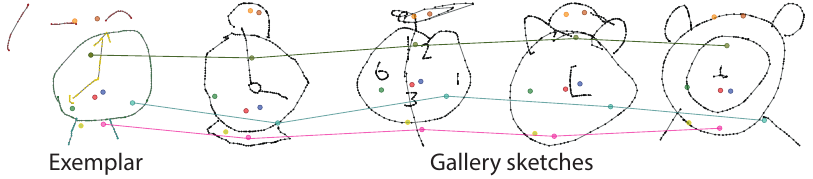}
\end{center}
\vspace{-6pt}
   \caption{Example keypoints predictions. For the visualization purposes we show only 9 keypoints. The exemplar lines trace the position of a keypoint across sketches.}
\label{fig:keypoints}
\vspace{-10pt}
\end{figure}

\subsection{Sketch-level transformation estimation}
We formulate the global sketch-level alignment task as a shape-matching problem \cite{sorkine2009least}, that allows \new{us} to analytically compute a rigid transformation between the two sets of corresponding points: 
\begin{equation}
\label{global_eq}
(\hat{R}, \hat{t})  = \underset{R,t}{\operatorname{argmin}} \frac{1}{M} \sum_{j = 1}^{M} \parallel (R x_i + t) - e_{i} \parallel ^ {2}, 
\end{equation}
where $M$ is the total number of point pairs, $x_i$ and $e_j$ are the points in the target and exemplar sketches.

To obtain point correspondences between the two sketches, we adopt the recent method by Chen et al.~\cite{chen2020unsupervised}, which estimates semantically consistent sets of keypoints in unsupervised manner. 
We first obtain feature representation for each sketch point\old{,} using the \old{encode we describe}\new{encoder described} in Sec.~\ref{sec:sketch_representation}.
The shared multi-layer percepetron (MLP) $\Omega_{\theta_1}$ is then trained to predict for each point $v_i$ a probability map $P = \{p_1,p_2,\ldots,p_M\}$\old{.}\new{, where} each element $p^i_j$ represents the probability of the point $v_i$ to be a  $j$-th keypoint\old{ $k_j$}. 
\old{The}\new{Thus, the} keypoints are \old{thus }computed as $k_j = \sum_{i=1}^L v_ip_j^i$. 
Note that the \old{extracted}\new{predicted} keypoints are not selected from the sketch stroke points\new{,} \old{but}\new{however they} are \old{en}forced to lie close to the input strokes by defining the Chamfer loss between the \old{the }input points $V \subset  \mathbb{R}^{2 \times N} $ and predicted keypoints $K_V  \subset  \mathbb{R}^{2 \times M} $, which we denote as $\mathcal{L}_{CD}(V, K_V)$. 
In Fig.~\ref{fig:keypoints} we demonstrate example extracted keypoints. Since we search for the transformation optimal in a least square sense the method is tolerant to small imprecision in the keypoints locations.

\subsection{Stroke-level transformation estimation}
Given an aligned target sketch $\hat{X}$ and an exemplar sketch $E$, we next predict stroke-level transformations $(R_j,\mathbf{t}_j,\mathbf{\sigma}_j)$ for each sketch stroke $\mathbf{s_j} \in E$. 
To predict stroke level transformation we use a network $\Phi_{\theta_2}$ consisting of successive multi-layer perceptrons (MLPs), followed by a ReLu activation function. The stroke transformation is calculated as
\begin{equation}
(R_j,\mathbf{t_j},\sigma_j) = \Phi_{\theta_2}([\mathcal{F}^{sketch}_{\hat{X}}, \mathcal{F}^{sketch}_{E}, \mathcal{F}^{stroke}_{\mathbf{s}_j \in E} ])
\label{eq:stroke_eq}
\end{equation}
where [*,*] denotes the vector concatenation operation, and $\mathcal{F}^{sketch}_{\hat{X}}$, $\mathcal{F}^{sketch}_{E}$ are the global embedding vectors of a globally aligned target sketch $\hat{X}$ and an exemplar $E$; $\mathcal{F}^{stroke}_{\mathbf{s}_j}$ is an embedding vector of the j-th stroke of the exemplar $E$.

To train this deformation we compute the Chamfer loss between the deformed exemplar $\hat{E}$, obtained by substituting the prediction result in Eq.~\ref{eq:stroke_eq} to Eq.~\ref{eq:deformed_template}, and the globally aligned target sketch $\hat{X}$: $\mathcal{L}_{CD}(\hat{E}_{\hat{X}}, \hat{X})$. We also compute the mean square error (MSE) distance between their keypoints: $\mathcal{L}_{MSE}(K_{\hat{E}_{\hat{X}}}, K_{\hat{X}})$. 

We assume that sketch-level deformation compensates for the large distances between structure points of the exemplar and target sketches, and only small deformations are required on a stroke level. 
During training we constrain each rotation matrix $R_j \subset  \mathbb{R}^{2 \times 2} $ to be close to an orthogonal matrix: $\mathcal{L}_{orth,j} = \Vert I - R_j R_j^T \Vert^2$.
Then, since the rotation matrix can be expressed in terms of one angle $\alpha_j$, we use a soft constraint on the stroke rotation to not exceed 30 degrees.  We achieve this by constraining each of the matrix elements to lie in the respective range. For instance, we constraint the first row and column element $r^j_{11}$ of the rotation matrix $R_j$, which encodes $\cos(\alpha_j)$ to lie in the interval $[\frac{\sqrt{3}}{2}, 1]$:
\begin{equation}
     \mathcal{L}_{rot,j}(r_{11}) =   \max(0, \frac{\sqrt{3}}{2} - r^j_{11} ) + \max(0, r^j_{11} - 1.0).  
\end{equation}

Similarly, we use a soft constraint on the stroke scaling $\mathbf{\sigma}=[\sigma^x, \sigma^y] \subset  \mathbb{R}^{2} $ to belong to the interval $[0.5, 2.0]$\old{with}\new{, we denote this loss as} $\mathcal{L}_{\sigma,j}$.
Finally, we constraint strokes translation vectors $\mathbf{t}_j$ to have a small norm:
$\mathcal{L}_{t,j} = \Vert  \mathbf{t}_j \Vert$.

Constraining stroke\new{-}level deformations\old{,} and exploiting keypoints\old{,} allows us to avoid erroneous deformations\old{,} and ensures that the deformations comply with the two rules listed in Sec.~\ref{sec:deformation}.

The full loss used to train stroke-level deformations is defined as
\begin{align}
\begin{split}
 \mathcal{L}_{strokes} &= \beta \mathcal{L}_{CD}(\hat{E}_{\hat{X}}, \hat{X}) + \gamma \mathcal{L}_{MSE}(K_{\hat{E}_{\hat{X}}}, K_{\hat{X}}) \\
  & + \gamma \frac{1}{\new{|E|}}\sum_j^{\new{|E|}} \left( \mathcal{L}_{orth,j} +  \mathcal{L}_{rot,j} +  \mathcal{L}_{\sigma,j} + \mathcal{L}_{t,j} \right),
\end{split}
\label{eq:L_strokes}
\end{align}
where \old{$\#E$}\new{$|E|$} is the number of strokes in the exemplar sketch.


\subsection{Sketch Encoding}
\label{sec:sketch_representation}

We exploit \new{a} graph convolutional network (GCN) to obtain sketch\old{-level}, stroke\old{-level}, and point level features. 
The network architecture we use is similar to the architecture of the global branch used in \cite{yang2020sketchgcn}, \old{building}\new{which is based} on the study of GCNs by Li et al.~\cite{li2019deepgcns}. 
We found a single branch to perform better than the full architecture proposed in \cite{yang2020sketchgcn}. 

The network consists of 4 layers with residual connections \cite{li2019deepgcns}. At each layer we construct a sketch graph $\mathcal{G} = (\mathcal{V},\mathcal{E})$, where $\mathcal{V} = V$ is a set of all sketch points, while $\mathcal{E}$ consists of two types of edges: static and dynamic. 

Dynamic edges result in large receptive field and improved performance\old{,} when combined with residual connections \cite{li2019deepgcns}. 
We construct dynamic edges using a Dilated k-NN strategy proposed in \cite{li2019deepgcns}. It first selects $k\times d$ nearest neighbors at each layer, and then constructs edges by selecting every $d$-th neighbor. We also implement the stochastic dilation, which\old{,} with \old{a }probability $\epsilon = 0.2$\old{,} \new{instead} selects $d$ neighbors uniformly from $k\times d$ nearest neighbors\old{, instead}.
Following \cite{yang2020sketchgcn}, we select $k = 4$, and set the dilation rate $d$ to 1, 4, 8, 16 for successive layers. 

Since points connectivity in strokes encodes\old{ an} important information about a sketch, similar to \cite{yang2020sketchgcn}, we combine\old{ the} dynamically constructed edges with static edges that are obtained by connecting consequent stroke\old{s} points. 

We use the convolutional operation, proposed in \cite{wang2019dynamic}, to extract point-level features $\mathcal{F}^{point}_{i}$. 
Then, the stroke\new{-}level features are define\old{s}\new{d} as 
\begin{equation}
\mathcal{F}^{stroke}_{s_j\in V} = \max_{i: v_i \in s_j } \mathcal{F}^{point}_{i}, 
\end{equation}
where $s_j$ denotes $j$-th strokes of a sketch with a point set $V$, and $v_i$ are all \old{the }points belonging to the stroke $s_j$. 
Similarly, the sketch embedding vector is defined as
\begin{equation}
\mathcal{F}^{sketch}_V = \max_{i: v_i \in V } \mathcal{F}^{point}_{i}, 
\end{equation}
where \old{a}\new{the} max-pooling is performed over all sketch points.

\subsection{Segmentation}
\label{sec:segmentation}
To obtain a label for each segment\new{,} we train \old{a}\new{the} label probability function $\tau$, which takes as \old{an }input a sketch point coordinate $v_i = (v_i^x,v_i^y)$, a sketch embedding vector  $\mathcal{F}^{sketch}_V$, and a stroke embedding vector  $\mathcal{F}^{stroke}_{s_j\in V: v_i \in s_j}$.
We condition the predictor on the stroke, since the points which belong to the same stroke are likely to have the same label.
The label probability function consists of the successive MLP layers with ReLu activation functions:
\begin{align}
    \tau(v_i) \coloneqq \tau_{\theta_3}(v_i, \mathcal{F}^{stroke}_{s_j\in V: v_i \in s_j}, \mathcal{F}^{sketch}_V) ,
\label{eq:segment}
\end{align}
such that $\tau: \mathbb{R}^{1\times (2+2K)} \rightarrow [0,1]^L$, where $L$ is a number of labels, and $K = 128$ is a length of sketch/stroke embedding vectors.

\old{At training time}\new{During training}, for each $\hat{E}_{\hat{X_t}}$, where $t$ goes over all \old{the }target sketches in a batch, we compute the cross entropy classification loss at each point $\hat{e}_{i,\hat{X_t}} \in \hat{E}_{\hat{X_t}}$\old{ with a label $l_{it}$}\new{, where we denote its ground-truth label as $l_{it}$}:
\vspace{-4pt}
\begin{equation}
    \mathcal{L}_{CE} = \sum_{i=1}^{N}\mathcal{L}_{\text{cross entropy}}(
			\tau_{\theta_3}(\hat{e}_{i,\hat{X_t}}), l_{it}).
\end{equation}

\subsection{Training and Losses}

We train our segmentation network in the end-to-end manner, where the keypoints prediction network $\Omega_{\theta_1}$, the stroke morphing module $\Phi_{\theta_2}$, and the label probability distribution function $\tau_{\theta_3}$ are trained jointly. The full loss is defined as
\begin{align}
\begin{split}
\mathcal{L} & = \alpha  \frac{1}{\mathcal{\vert B \vert }}\sum_{t=1}^{\mathcal{\vert B \vert }} \mathcal{L}_{CD}(X_t, K_{X_t})  + \\
 & \frac{1}{\mathcal{\vert B \vert }}\sum_{t=1}^{\mathcal{\vert B \vert }} \left( \mathcal{L}_{strokes}(\hat{E}_{\hat{X_t}}, \hat{X_t}) + \delta \mathcal{L}_{CE} (\hat{E}_{\hat{X_t}}) \right),
\end{split}
\label{eq:L_full}
\end{align}
where $\vert B \vert $ is the number of sketches in each batch.

\section{Experiments}

\subsection{Datasets}
We evaluate our method on four sketch datasets: SPG \cite{li2018universal},  Huang14 \cite{huang2014data}, TU-Berlin \cite{eitz2012humans, li2018fast} and creative birds \cite{ge2020creative}. 
The SPG dataset consists of 25 categories with 800 sketches each, annotated with stroke-level semantic labels. The sketches come from the QuickDraw dataset \cite{ha2017neural} -- a vector sketches dataset,  collected via an online game where the players are asked to draw objects within 20 seconds. 
Annotated TU-Berlin dataset \cite{li2018fast} contains 5 categories with 80 sketches each from the TU-Berlin dataset \cite{eitz2012humans}. The participants were asked to produce a sketch of a given category within 30 minutes window. The labeling is obtained through crowd-sourcing. Huang14 dataset \cite{huang2014data} consists of 10 categories with 30 sketches each, drawn by 3 participants from \old{a }reference photo\new{s}. 
Creative birds and creatures \cite{ge2020creative} is a challenging dataset, featuring non-canonical representation of birds and arbitrary creatures. 
The participants are provided with a first stroke\old{,} and are asked to place an eye where they like\old{,} and to visualize how the stroke and the eye can be incorporated into a creative sketch, \eg, of a bird. 
We evaluate on the creative birds only. 
The segmentation labels vary significantly among people on the creative creatures, and pose a poor ground-truth for our studies.
                                                                                                                                                                                                                                                                                                                                               
\subsection{Alternative solutions} 
\label{sec:baselines}
As discussed in Sec.~\ref{sec:related_work}, we are the first to consider the problem of one/few shot sketch segmentation. 
\new{Therefore, we}\old{We therefore} compare with two state-of-the-art supervised \new{methods: the} semantic sketch segmentation method \cite{yang2020sketchgcn} and \new{the} perceptual grouping method \cite{li2018universal}\old{,} trained with one or a few example sketches, and \new{existing} few shot segmentation methods for 3D shapes \cite{wang2020few, chen2020unsupervised}.
\textbf{SPGG} \cite{li2018universal} exploits a sequence-to-sequence variational autoencoder to obtain sketch encoding\old{,} and aims at globally consistent segments grouping.
This method does not predict a label for the group, \new{therefore,} for the evaluation we assign each group \old{the}\new{a} label\old{,} based on the overlap with the ground-truth grouping. 
\textbf{SGCN} \cite{yang2020sketchgcn} encodes a sketch with a GCN, consisting of two branches with static and dynamic convolutions. It is trained with the cross-entropy loss. This work gives the state-of-art segmentation results under supervised segmentation setting. In our work\new{,} we use a similar sketch encoding architecture, it thus is a strong baseline for our method. 
\textbf{FLSS} \cite{wang2020few} serves as \old{a}\new{the} main baseline for our method\old{,} and addresses a few-shot 3D shape segmentation. Unlike us, it assumes that all shapes have similar global shape alignment, and models morphing from the exemplar annotated 3D shape to the target shape by predicting per point offset vectors.
\textbf{ISPP} \cite{chen2020unsupervised} is an encoder-decoder based architecture for semantically meaningful keypoints selection on a 3D point cloud. In our work we exploit this architecture to supervise sketch morphing. 
To perform a label transfer directly, we\old{,} first{,} \new{find} for each point in the target sketch\old{, find} the closest keypoint in \new{the }Euclidean space. 
\old{Then, w}\new{W}e \new{then} select the closest keypoint from the exemplar sketch to the selected keypoint in the feature space and transfer its label. 
The original ISPP \cite{chen2020unsupervised} method relies on the PointNet++ encoder. 
In all our comparisons, we instead use our GCN encoder as it results in better performance (\old{we refer the reader to the supplemental material for the detailed evaluation}\new{we provide the detailed evaluation in Sec.~\ref{sec:ablation_studies}}).

\subsection{Implementation Details}
\label{sec:implemntation_details}

To obtain an $N$ point-set sketch representation, we first simplify the sketches with Ramer-Douglas-Peucker algorithm to nearly 256 points. Then, if there are less than 256 points, we use a simple sampling strategy of dividing random segments in the middle till we have roughly 256 points. 
In case if there are still more points\new{,} we randomly skip some points.
We use $M-256$ keypoints in our work. 
For \new{the} SPGG \new{method,} we used the original points sampling.
We set $\alpha = 1.0$, $\delta = 0.02$ in Eq.~\ref{eq:L_full}, $\beta = 0.2$ and $\gamma = 50$ in Eq.~\ref{eq:L_strokes}. 

For all methods\new{,} we perform data augmentation by rotating sketches by a randomly chosen angle from the interval $[-\frac{\pi}{12},\frac{\pi}{12}]$.
We use Adam optimizer ($\beta_1 =0.9, \beta_2=  0.999$) with a learning rate $5e-5$ and a batch size 24.
At inference, to obtain the labeling via Eq.~\ref{eq:segment}, we first estimate our hierarchical deformation, then the label of a point $v_i$ is obtained as follows $\tau(v_i) = \tau_{\theta_3}(v_i, \mathcal{F}^{stroke}_{s_j\in \hat{X}: v_i \in s_j}, \mathcal{F}^{sketch}_{\hat{E}})$.


\begin{table*}[t]
 \begin{center}
\footnotesize
\resizebox{\linewidth}{!}{%
\begin{tabular}{l|rr|rr|rr|rr|rr|rr|rr|rr|rr|rr}
                                    & \multicolumn{10}{c|}{\textbf{P-metric}}                                                                                                                                                                                                                                                                                                  & \multicolumn{10}{c}{C-metric}                                                                                                                                                                                                                                                                                                            \\ \cline{2-21} 
                                    & \multicolumn{2}{c|}{SPGG \cite{li2018universal}}                & \multicolumn{2}{c|}{SGCN \cite{yang2020sketchgcn}}               & \multicolumn{2}{c|}{FLSS \cite{wang2020few}}                    & \multicolumn{2}{c|}{ISPP \cite{chen2020unsupervised}}           & \multicolumn{2}{c|}{Ours}                                       & \multicolumn{2}{c|}{SPGG \cite{li2018universal}}                & \multicolumn{2}{c|}{SGCN \cite{yang2020sketchgcn}}               & \multicolumn{2}{c|}{FLSS \cite{wang2020few}}                     & \multicolumn{2}{c|}{ISPP \cite{chen2020unsupervised}}            & \multicolumn{2}{c}{Ours}                                       \\
\hline
\end{tabular}
}
 \caption{Numerical evaluation on the SPG dataset \cite{li2018universal}: first 25 categories; on the 'airplane' category from TUBerlin \cite{eitz2012humans} and Huang14 \cite{huang2014data} datasets; on creative birds \cite{ge2020creative}. $\mu$ denotes the average accuracy over 5 runs with 5 randomly chosen exemplars, and $\sigma$ is the standard deviation of the 5 runs results. The five categories in bold are the categories we use for the detailed analysis and ablation studies. `ref.' refers to the results refined by recomputing the label per point based on the dominant stroke label. 
 }
  \label{tab:1}
\end{center}
\end{table*}

\begin{figure*}[t]
\begin{center}
   \includegraphics[width=\textwidth]{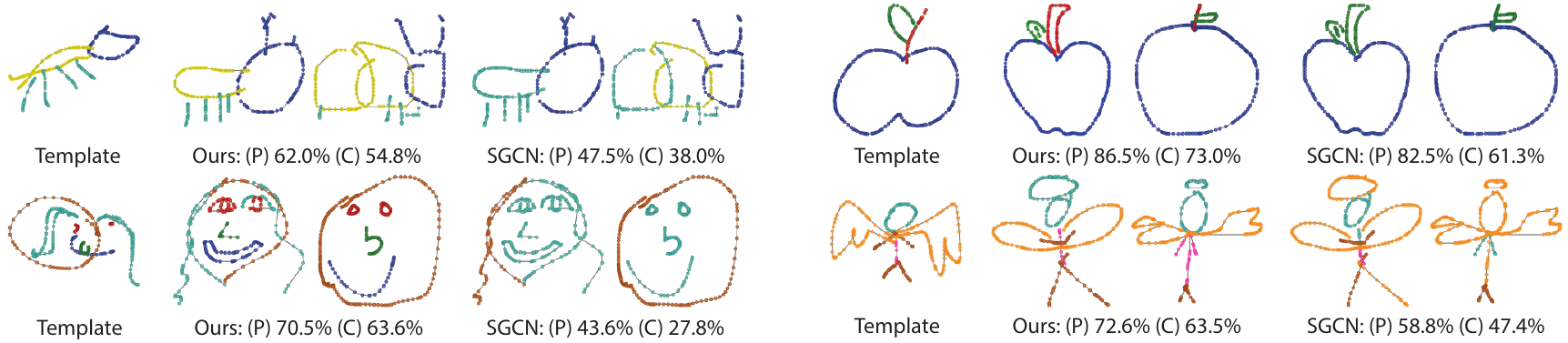}
\end{center}
   \caption{Comparison between our method and one-shot setting for SGCN \cite{yang2020sketchgcn}. Our method is generally more accurate. The most advantage can be observed on more abstract exemplars (\eg, face) or complex categories (\eg, angel). The numbers indicate (P) point and (C) component average accuracy over the category for the shown exemplar. The points show the sampled points, and the grey thin lines demonstrate their connectivity into strokes.}
\label{fig:SketchGCN_vs_ours}
\end{figure*}

\subsection{Evaluation}
\label{sec:evalaution}
We evaluate with traditional segmentation metrics: (a) \emph{pixel/point accuracy (P-metric)} -- the fraction of points that are assigned with a correct label and (b) \emph{component accuracy (C-metric)} -- the number of correctly labeled components divided by the total number of components.

It is challenging to develop a faithful evaluation of the few-shot segmentation\old{,} since each ground-truth has only one set of labels, which can have different granularity within the same category and dataset. 
There are two challenging cases: (1) The ground-truth labeling of the target is more fine-grained than the labeling in the exemplar: \eg, a user does not draw or mark a window for an airplane in the exemplar, while the target sketch ground-truth contains a window labeling; (2) The exemplar labeling is more fine-grained than the ground-truth labeling of the target: \eg, the user labels the plane tail in the exemplar, while the target sketch ground-truth considers it to be a body of the plane.
The first case we address automatically, by ignoring during the evaluation the points of the target sketch with the label not-existing in the exemplar. 
The second case is more challenging since it is impossible to automatically detect when the ground-truth labeling has a different labeling granularity.
In the supplemental\new{,} we provide a more restrictive evaluation on subsets of sketches that have the same set of labels as an exemplar. We though do not observe much differences between the two evaluation \old{methods}\new{approaches}.

\begin{figure*}[t]
\begin{center}
   \includegraphics[width=\textwidth]{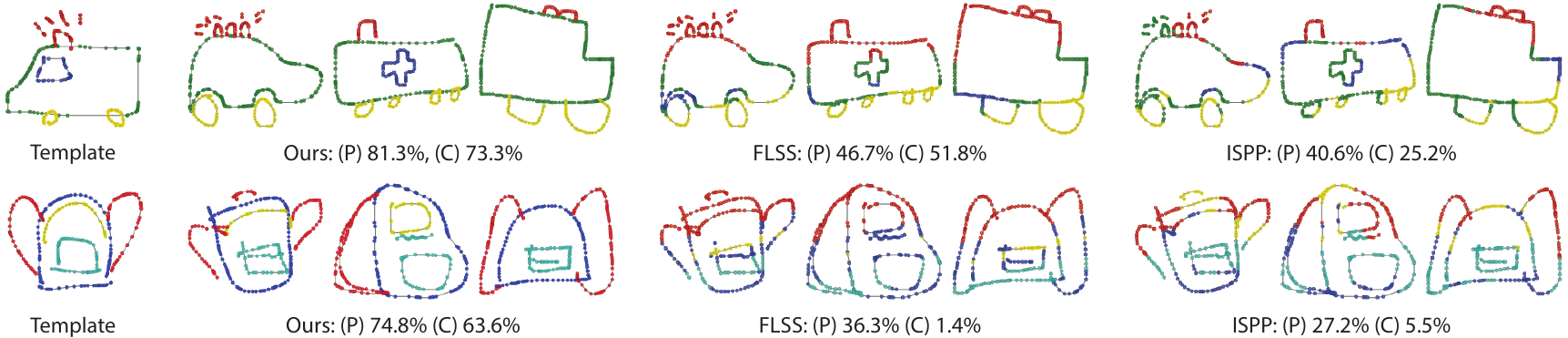}
\end{center}
   \caption{Comparison of our method with FLSS \cite{wang2020few} and ISPP \cite{chen2020unsupervised} on the example exemplars from `ambulance' and `backpack' categories of the SPG dataset \cite{li2018universal}.  The numbers indicate (P) point and (C) component average accuracy for the shown exemplar. The points show the sampled points, and the grey thin lines demonstrate their connectivity into strokes.}
\label{fig:FewShot_vs_ours}
\end{figure*}

\subsection{One-shot segmentation: Performance on average}
We first evaluate a one-shot segmentation scenario. 
For each category\new{,} we randomly select 5 sketches as exemplars\old{,} and report an average accuracy over\old{ the} 5 runs. 
We train on the training subset of \new{the} SPG dataset \cite{li2018universal} and evaluate on the test subset of \new{the} SPG dataset and the overlapping categories from \new{the} Huang14 \cite{huang2014data} and TU-Berlin \cite{eitz2012humans, li2018fast} datasets, demonstrating that \old{the}\new{our} segmentation \new{model} can generalize to sketches from different distributions and to sketches that are not observed during training. 
Since Huang14 and TU-Berlin have slightly different set\new{s} of labels, namely\new{,} there is no `window' label in both datasets, we change in all used exemplars the `window' label to the `body' label. 
Similarly, `airplane\_horistab' and `airplane\_vertstab'\old{,} labels are merged to a single `tail' label. 
We also train and test on the creative birds \cite{ge2020creative} dataset.

Tab.~\ref{tab:1} shows the numerical evaluation of all considered methods. 
Figures \ref{fig:SketchGCN_vs_ours}, \ref{fig:FewShot_vs_ours} show visual comparisons. 
Our method results in \new{the} highest point and component accuracy \old{on}\new{across} all \old{the }evaluated datasets and categories. 
On average on the SPG dataset\new{,} our method results in $10.8\%$ and $11.2\%$ higher point and component accuracy than the second best method SGCN \cite{yang2020sketchgcn}. 
\old{On}\new{In particular, for instance, on} the `duck' category\old{ it results in up to $28.4\%$ accuracy increase}\new{, the accuracy value for our method is higher by $28.4\%$}.
FLSS \cite{wang2020few} and ISPP \cite{chen2020unsupervised} have similar performance, \old{and}\new{while} FLSS performs slightly better. 
On average on the SPG dataset\new{,} our method results in $22.6\%$ and $37.1\%$ higher point and component accuracies values than FLSS.
On the `creative birds' dataset SGCN \cite{yang2020sketchgcn} performs the worst among all \old{the }methods, not being able to account for diversity in the dataset. 
ISPP \cite{chen2020unsupervised} method results in the second best performance \old{,} after our method\old{,} on this dataset.

Since the labeling in the considered datasets is defined per stroke (the strokes are broken into multiple at data-annotation stage if is needed), we additionally can perform an easy label refinement step\old{,} by assigning to a point a label dominant to the stroke the point belongs to (Tab.~\ref{tab:1} `ref.'). 
Note\old{ though}\new{, however,} that in sketches found in the wild\new{, a} \old{the }stroke can have several labels, and such \new{a} refinement step can reduce \old{the }segmentation accuracy. 
It can be seen that under this setting our method also performs the best. Only on the creative birds dataset our method gives \new{a} lower C-metric \new{value} than FLSS and ISPP, but still results in \new{a} higher P-metric \new{value}. 
On the SPG dataset FLSS and ISPP are still losing to SGCN and our methods, where our method gives $10.8\%$ and $11.3\%$ higher point and component accuracy \new{values} than SGCN.
After refinement, on the SPG dataset, the accuracy of our method increases just by $0.2/0.8 $ points on P/C-metrics, compared to by  $10.0/24.4$ points for ISSP and $8.9/19.7$ points for FLSS. 
Importantly, these results show that our method is able to directly \old{taking}\new{accounting for} points connectivity into strokes, but does not \new{im}pose a strict requirement of one label per stroke.

\subsection{One-shot segmentation: Robustness to number of parts, complexity and diversity}
For the remaining experiments, we select 5 categories from the \old{PSG}\new{SPG} dataset \cite{li2018universal} of varying complexity, by selecting categories with \new{a} different maximum number of parts: \emph{Apple} has at most three semantic parts, \emph{Duck} -- four, \emph{Ambulance} -- five, \emph{Face} -- seven, and \emph{Pig} -- eight.

We first evaluate the robustness of our method under different number of part labels\old{,} and compare to the second best method on the SPG dataset -- SGCN \cite{yang2020sketchgcn}. 
For each of the five categories\new{,} we randomly select \old{3 exemplars with}\new{exemplars from}
\setlength{\columnsep}{10pt}%
\begin{wrapfigure}[11]{l}{4.2 cm}
\centering
\vspace{-0pt}
\includegraphics[width=4.1cm]{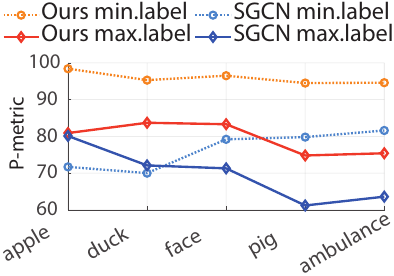}
   \caption{Performance based on number of parts.}
\label{fig:num_label}
\end{wrapfigure}
\new{the sketches containing the} minimum and maximum number of labels, \new{3 exemplars for each setting}. 
Fig.~\ref{fig:num_label} plots \old{an}\new{the} average accuracy over three runs\old{,} and shows that our method consistently outperforms SGCN.
Our method achieves an average \new{accuracy value of $95.9\%$} over the five classes \old{accuracy of $95.9\%$ }on a simpler task with little parts (yellow dashed line). 
Finally, our method is more robust on complex sketches: as the sketches complexity and diversity increase from `apple' to `pig' categories under the challenging task with many labels, SGCN performance degrades by $18.9$ points (solid blue line)\old{, versus} \new{compared to} just $6.1$ points with our method (solid red line).

\subsection{One-shot vs. few-shot}
\label{sec:one_vs_few}
Tab.~\ref{tab:one_vs_few} shows that the performance improves if there are several \old{templates}\new{exemplars} available\old{,} and our results consistently outperform \new{those of} SGCN. 
For \old{these}\new{this} experiment\new{,} we select 5 exemplars for each category with consistent semantic parts, selecting the most frequent number of parts in each class ground-truth labeling.  
For the results on 1 exemplar, we average the labeling accuracy over labeling results with each exemplar individually. 
For the 3 exemplars, we average over all unique subsets of 5 exemplars. 
When multiple exemplars are provided, to choose the best we perform the hierarchical deformation estimation and select the one that results in the smallest Chamfer distance. 

\begin{table}[h]
 \begin{center}
\footnotesize
\resizebox{\linewidth}{!}{%
\begin{tabular}{ll|cr|cr|cr}
                                                   & \textbf{}         & \multicolumn{2}{c|}{\textbf{1 exemplar}}                     & \multicolumn{2}{c|}{\textbf{3 exemplars}}                    & \multicolumn{2}{c}{\textbf{5 exemplars}}                   \\ \hline
\multicolumn{1}{l|}{}                              & \textbf{Category} & (P)                               & \multicolumn{1}{c|}{(C)} & (P)                               & \multicolumn{1}{c|}{(C)} & (P)                               & \multicolumn{1}{c}{(C)} \\ \hline
\multicolumn{1}{c|}{Ours}                          & Average 5         & \multicolumn{1}{r}{\textbf{85.7}} & \textbf{76.0}            & \multicolumn{1}{r}{\textbf{91.3}} & \textbf{85.4}            & \multicolumn{1}{r}{\textbf{92.4}} & \textbf{87.8}           \\ \hline
\multicolumn{1}{c|}{SGCN \cite{yang2020sketchgcn}} & Average 5         & \multicolumn{1}{r}{77.9}          & 63.4                     & \multicolumn{1}{r}{86.5}          & 79.4                     & \multicolumn{1}{r}{86.5}          & 79.2  \\ \hline                 
\end{tabular}
}
 \caption{One shot vs.~few shot. See Sec.\ref{sec:one_vs_few} for the details.
 }
  \label{tab:one_vs_few}
    \end{center}
    \vspace{-15pt}
\end{table}

\subsection{Generalization to unseen sketches of the same category}

All results in our work are evaluated on sketches that are not observed during training. Yet, the target unlabeled sketches in practice can be used for training. Our results in Table \ref{tab:generalization} demonstrate that the performance is the same whether the target sketches are used for training directly or not, showing \new{the} good generalization \new{property} of our method. 

\begin{table}[]
\centering
\small
\begin{tabular}{cc|cc|cccc}
\multicolumn{4}{c|}{\textbf{P-metric}}                                                                    & \multicolumn{4}{c}{\textbf{C-metric}}                                                                          \\ \hline
\multicolumn{2}{c|}{\textbf{Training}}                 & \multicolumn{2}{c|}{\textbf{Test}}                  & \multicolumn{2}{c}{\textbf{Training}}                       & \multicolumn{2}{c}{\textbf{Test}}                  \\ \hline
$\mu$                    & $\sigma$                 & $\mu$                    & $\sigma$                 & $\mu$                    & \multicolumn{1}{c|}{$\sigma$} & $\mu$                    & $\sigma$                 \\ \hline
\multicolumn{1}{r}{85.9} & \multicolumn{1}{r|}{6.7} & \multicolumn{1}{r}{86.2} & \multicolumn{1}{r|}{6.9} & \multicolumn{1}{r}{77.7} & \multicolumn{1}{r|}{10.3}     & \multicolumn{1}{r}{77.8} & \multicolumn{1}{r}{11.1} \\ \hline
\end{tabular}%
\caption{Comparison of the segmentation results on the unlabeled sketches from the training set (`Training') and on unseen sketches (`Test'). The results are averaged over the five selected categories and over the runs with the same exemplars as in Tab.~\ref{tab:1}.}
\label{tab:generalization}
\end{table}

\begin{figure}[t]
\begin{center}
   \includegraphics[width=1.0\linewidth]{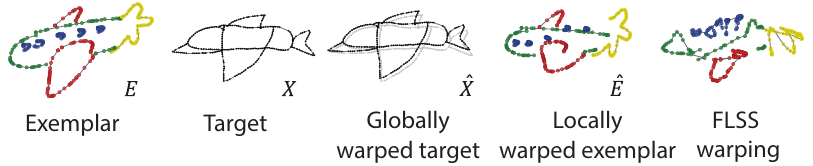}
\end{center}
   \caption{The deformation process: our method vs. FLSS \cite{wang2020few}.}
\label{fig:deform}
\end{figure}

\begin{table}[t]
\centering\captionsetup{font=footnotesize}
\resizebox{\linewidth}{!}{%
\begin{tabular}{cc|cc|cc|cc|cc}
\multicolumn{2}{c|}{\textbf{Ours full}}              & \multicolumn{2}{c|}{\textbf{No G.A.}}                 & \multicolumn{2}{c|}{\textbf{Chamfer G.A.}}  & \multicolumn{2}{c|}{\textbf{Reverse G.A.}} & \multicolumn{2}{c}{\textbf{No  $\mathcal{F}^{stroke}$}} \\ \hline 
\textbf{(P)}             & \textbf{(C)}             & 
\textbf{(P)}             & \textbf{(C)}             & 
\textbf{(P)}             & \textbf{(C)}             & 
\multicolumn{1}{c}{(P)}         & \textbf{(C)}   	& 
\multicolumn{1}{c}{(P)}         & \textbf{(C)}       \\ \hline
\multicolumn{1}{r}{86.2} & \multicolumn{1}{r|}{77.8} & 
\multicolumn{1}{r}{68.5} & \multicolumn{1}{r|}{60.1} & 
\multicolumn{1}{r}{49.6} & \multicolumn{1}{r|}{29.1} & 
\multicolumn{1}{r}{57.9} & \multicolumn{1}{r|}{44.1} &
\multicolumn{1}{r}{60.8} & \multicolumn{1}{r}{36.0} 
 \\ \hline     
\end{tabular}

}
\vspace{1pt}
\caption{Ablation studies: The results are averaged over the five selected categories, over the runs with the same exemplars as in Tab.~\ref{tab:1}. `\textbf{No G.A.}': We skip the step of global transformation (G.A.), and perform only stroke-level deformations. `\textbf{Chamfer G.A.}': The global transformation is estimated with the Chamfer distance, instead of relying on the distances between keypoints. `\textbf{Reverse G.A.}': We estimate the global transformation from an exemplar towards a target. `\textbf{No  $\mathcal{F}^{stroke}$}': We remove $\mathcal{F}^{stroke}$ in Eq.~\ref{eq:segment}.  }
\label{tab:keypoints_vs_chamfer} 
\end{table}

\begin{table}[t]
\centering
\small
\begin{tabular}{cc|cc|cccc}
\multicolumn{4}{c|}{{\color[HTML]{000000} \textbf{P-metric}}}                                                                                                       & \multicolumn{4}{c}{{\color[HTML]{000000} \textbf{C-metric}}}                                                                                                                             \\ \hline
\multicolumn{2}{c}{{\color[HTML]{000000} \textbf{PointNet++}}}                   & \multicolumn{2}{c|}{{\color[HTML]{000000} \textbf{Ours}}}                        & \multicolumn{2}{c}{{\color[HTML]{000000} \textbf{PointNet++}}}                                        & \multicolumn{2}{c}{{\color[HTML]{000000} \textbf{Ours}}}                        \\ \hline
{\color[HTML]{000000} \textbf{$\mu$}} & {\color[HTML]{000000} \textbf{$\sigma$}} & {\color[HTML]{000000} \textbf{$\mu$}} & {\color[HTML]{000000} \textbf{$\sigma$}} & {\color[HTML]{000000} \textbf{$\mu$}} & \multicolumn{1}{c|}{{\color[HTML]{000000} \textbf{$\sigma$}}} & {\color[HTML]{000000} \textbf{$\mu$}} & {\color[HTML]{000000} \textbf{$\sigma$}} \\ \hline
\multicolumn{1}{r}{39.9}              & \multicolumn{1}{r|}{9.7}                 & \multicolumn{1}{r}{86.2}              & \multicolumn{1}{r|}{6.9}                 & \multicolumn{1}{r}{17.2}              & \multicolumn{1}{r|}{6.6}                                      & \multicolumn{1}{r}{77.8}              & \multicolumn{1}{r}{11.1}         \\       
\hline
\end{tabular}%
\caption{PointNet++ encoder \cite{qi2017pointnet++} vs. our GCN encoder. The results are averaged over the five selected categories, over the runs with the same exemplars as in Tab.~\ref{tab:1}.}
\label{tab:pointnet_encoder}
\end{table}

\begin{table}[t]
\centering
\resizebox{\linewidth}{!}{%
\begin{tabular}{cc|cc|cc|cc|cc}
\multicolumn{2}{c|}{\textbf{\begin{tabular}[c]{@{}c@{}}No  $\mathcal{L}_{rot}$, \\ $\mathcal{L}_{\sigma}$, $\mathcal{L}_{t}$\end{tabular}}} & \multicolumn{2}{c|}{\textbf{No  $\mathcal{L}_{rot}$}}   & \multicolumn{2}{c|}{\textbf{No $\mathcal{L}_{\sigma}$}} & \multicolumn{2}{c|}{\textbf{No $\mathcal{L}_{t}$}} & \multicolumn{2}{c}{\textbf{\begin{tabular}[c]{@{}c@{}}Ours\\ full\end{tabular}}} \\ \hline
\textbf{(P)}                                                        & \textbf{(C)}                                                          & \textbf{(P)}              & \textbf{(C)}               & \textbf{(P)}               & \textbf{(C)}               & \textbf{(P)}               & \textbf{(C)}               & \textbf{(P)}                            & \textbf{(C)}                           \\ \hline
\multicolumn{1}{r}{61.4}                                            & \multicolumn{1}{r|}{42.96}                                            & \multicolumn{1}{r}{81.04} & \multicolumn{1}{r|}{69.62} & \multicolumn{1}{r}{77.36}  & \multicolumn{1}{r|}{67.82} & \multicolumn{1}{r}{63.52}  & \multicolumn{1}{r|}{44.14} & \multicolumn{1}{r}{86.18}               & \multicolumn{1}{r}{77.84} \\
\hline            
\end{tabular}%
}
\caption{The role of constraints on stroke-level deformation. The results are averaged over the five selected categories, over the runs with the same exemplars as in Tab.~\ref{tab:1}. }
\label{tab:cosntraints}
\end{table}

\subsection{Discussion} Our first advantage over SGCN lies in the ability to generate structural variations on the exemplar, making the training robust towards different sketch abstractions and styles. Second, our global deformation step from the target to the exemplar lifts the requirement on the segmentation module to be rotation/reflection invariant. Our advantage over FLSS and ISSP methods lies in the ability to account for points connectivity into strokes, resulting in more meaningful labeling. Compared to ISSP, our robust segmentation network makes our algorithm more tolerant towards mistakes in the keypoints prediction. Finally, our sketch dedicated deformation model allows to better preserve \new{the} sketch structure than FLSS, as shown in Fig.~\ref{fig:deform}.

Our one-shot segmentation performance does not yet reach the performance of the fully supervised methods. For instance, the state-of-the-art SGCN under the fully supervised non-personalized (all sketches have different segmentation granularity) setting achieves accuracy of $96.8\%/94.3\%$ on P/C metrics. Nevertheless, as it can be seen in Sec.~\ref{sec:one_vs_few} the performance of the personalized sketch segmentation quickly increases when there are more exemplars available. 
Moreover, our solution consistently achieves higher accuracy in one-shot and few-shot personalized segmentation scenarios than the state-of-the-art segmentation method, demonstrating the efficiency of our deformation module.

\begin{figure*}[t]
\begin{center}
\includegraphics[width=1.02\linewidth]{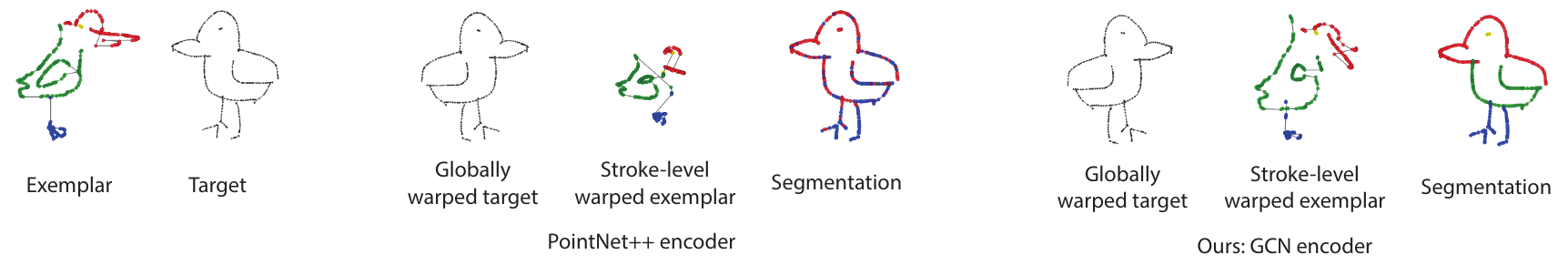}
\end{center}
\vspace{-10pt}
   \caption{Visual comparison of the deformation and segmentation results, depending on the used sketch encoder: PointNet++ \cite{qi2017pointnet++} or our GCN encoder.}
\label{fig:pointnet_encoder}
\vspace{-8pt}
\end{figure*}

\subsection{Ablation studies}
\label{sec:ablation_studies}

\paragraph{Global alignment (G.A.) and keypoints loss}

Tab.~\ref{tab:keypoints_vs_chamfer} (`Ours' vs.~`No G.A.') shows the importance of hierarchical estimation: If we remove the step of global alignment\new{,} the average over the 5 representative categories drops by $17.7$ points in terms of \new{both}\old{ the} point- and component-based accuracy. 
Similarly, we demonstrate the importance of globally warping target sketches towards an exemplar, rather than an exemplar towards target sketches  (`Ours' vs.~`Reverse G.A.'). 

Further, Tab.~\ref{tab:keypoints_vs_chamfer} (`Ours' vs.~`Chamfer G.A.') shows the importance of relying on keypoints instead of the Chamfer distance between the points of two sparse sketches. If the Chamfer distance is used, the alignment does not necessary respect the semantics of strokes.


\paragraph{Stroke-level information for segmentation}
We show that accounting for stroke-level information in the segmentation module (Sec.\ref{sec:segmentation}, Eq.~\ref{eq:stroke_eq}) has \new{a} high impact on the accuracy of the prediction (Tab.~\ref{tab:keypoints_vs_chamfer} (`Ours' vs.~`No  $\mathcal{F}^{stroke}$')).

\paragraph{Graph-based vs. point cloud-based encoder}

We demonstrate an advantage of a Graph Convolutional Network (GCN) over point cloud encoders for the few shot sketch segmentation problem. We exploit here the PointNet++ \cite{qi2017pointnet++} encoder, which we use instead of \new{the} GCN, keeping the architecture otherwise the same. 
Table \ref{tab:pointnet_encoder} and Fig.~\ref{fig:pointnet_encoder} show that \new{the} point cloud encoder is not capable of capturing well stroke-level information, resulting in poor segmentation performance.

\paragraph{Soft constraints on stroke-level deformation}
To evaluate the role of our soft constraints on stroke-level deformation, we first disable all the constraints $\mathcal{L}_{rot}$, $\mathcal{L}_{\sigma,j}$, $\mathcal{L}_{t,j}$, apart from $\mathcal{L}_{orth}$ in Eq.~\ref{eq:L_strokes}. 
We then remove each of the three terms individually. 
Table \ref{tab:cosntraints} demonstrates the importance of these constraints. $\mathcal{L}_{t}$ is the most important since it limits how far strokes can move from their original positions, ensuring global sketch structure maintenance.

Despite such constraints help to preserve the relative part relationship, they limit the space of achievable deformations. For instance, Fig.~\ref{fig:deform} shows that our model just roughly aligns the exemplar to the target. To improve on our results, future work should investigate alternative deformation models that can preserve the relative part relationship while achieving better alignment with the target sketch.

%

\section{Conclusion}

We present the first one-shot personalized sketch segmentation method and study a set of alternative\old{s} solutions \new{constructed by} adopting the state-of-the-art segmentation and perceptual grouper networks, and two 3D shape few-shot segmentation networks. 
We address this problem by estimating the deformation from an exemplar sketch towards a target sketch and training a robust part label predictor network on the warped exemplars. Our key contributions lie in proposing a hierarchical deformation model that works at both sketch- and stroke-level. 
Our hierarchical \old{two-ways}\new{bi-directional} deformation model allows \new{us} to explicitly account for ambiguity in global sketch orientation, resulting in more robust segmentation results. 
We also demonstrate the importance of taking stroke connectivity into consideration and compare point cloud and graph-based encoders.
We show that our method by far outperforms all existing alternatives, showing robust performance on the highly abstract exemplars and complex categories.


%





\ifCLASSOPTIONcaptionsoff
  \newpage
\fi



\bibliographystyle{IEEEtran}
\bibliography{egbib}
%



%
\begin{IEEEbiography}[{\includegraphics[width=1in,height=1.25in,clip]{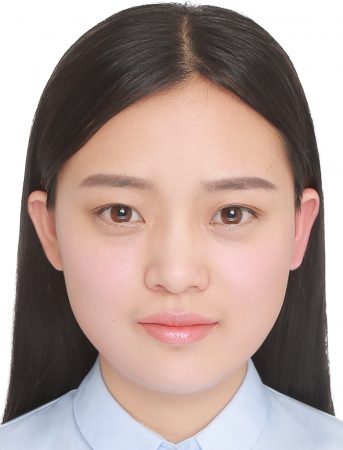}}]{Anran Qi}
is a PhD student of SketchX Lab, Centre for Vision Speech and Signal Processing (CVSSP), University of Surrey. Her sketch focus on sketch oriented or aided 3D shaped research topic, including sketch-based 3D shape retrieval and 3D shape recustruction.
\end{IEEEbiography}

\begin{IEEEbiography}[{\includegraphics[width=1in,height=1.25in,clip]{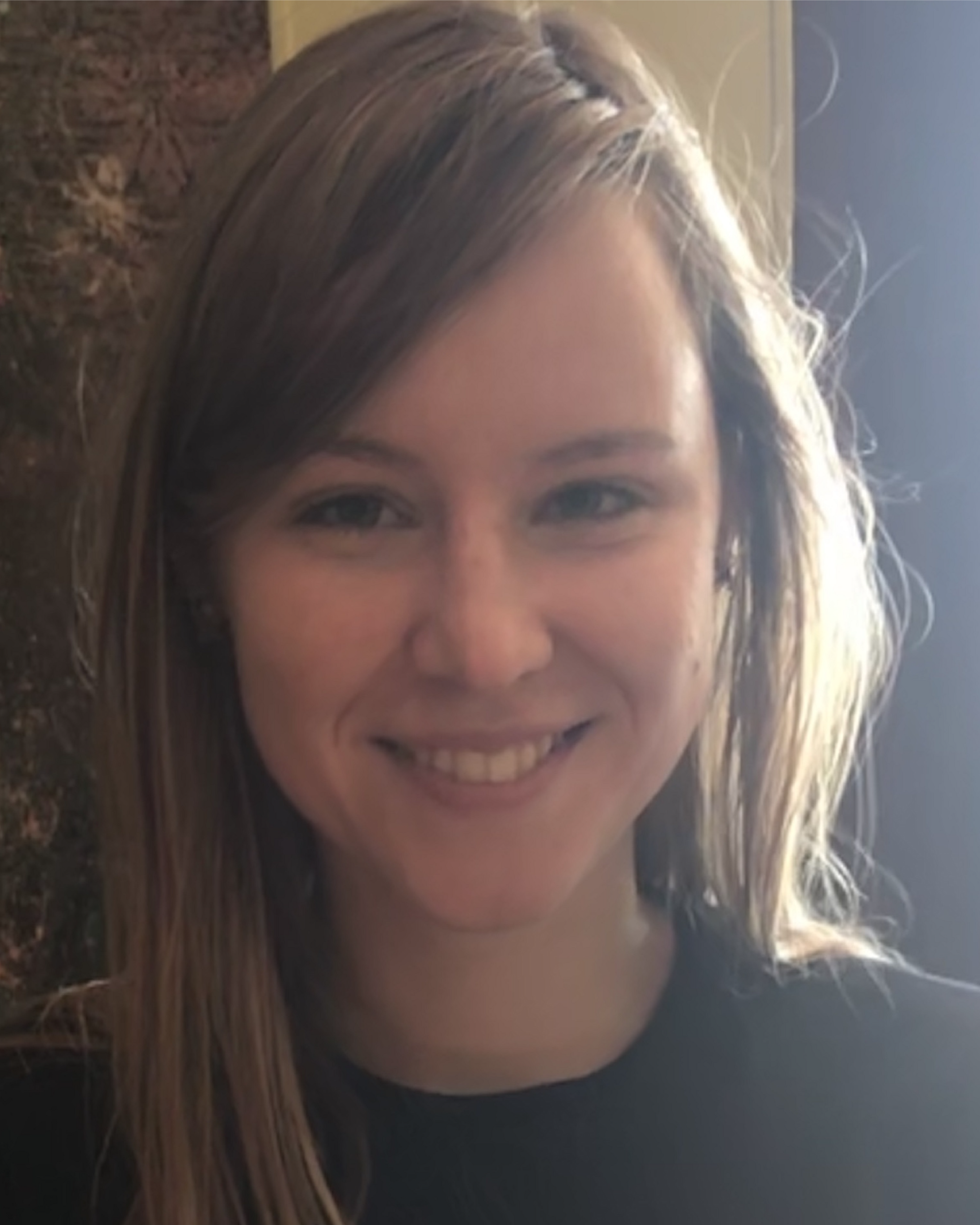}}]{Yulia Gryaditskaya} 
%
is a Lecturer in Artificial Intelligence at the Centre for Vision Speech and Signal Processing (CVSSP) and Surrey Institute for People-Centred AI.
Previously, she was a postdoctoral researcher at Inria, Sophia Antipolis, France. 
She obtained her PhD in 2017 on Computer Vision and Graphics from the Max-Planck institute for informatics, Saarbruecken, Germany. She got her diploma from Lomonosov Moscow State University, Moscow, Russia.
Her research interests cover sketch-based modeling, sketch beautification, geometric deep learning, sketch classification, sketch generation, high dynamic range image and video capture, tone-mapping and calibration, depth estimation from the structured light-fields, materials representation and editing.
\end{IEEEbiography}

\begin{IEEEbiography}[{\includegraphics[width=1in,height=1.25in,clip]{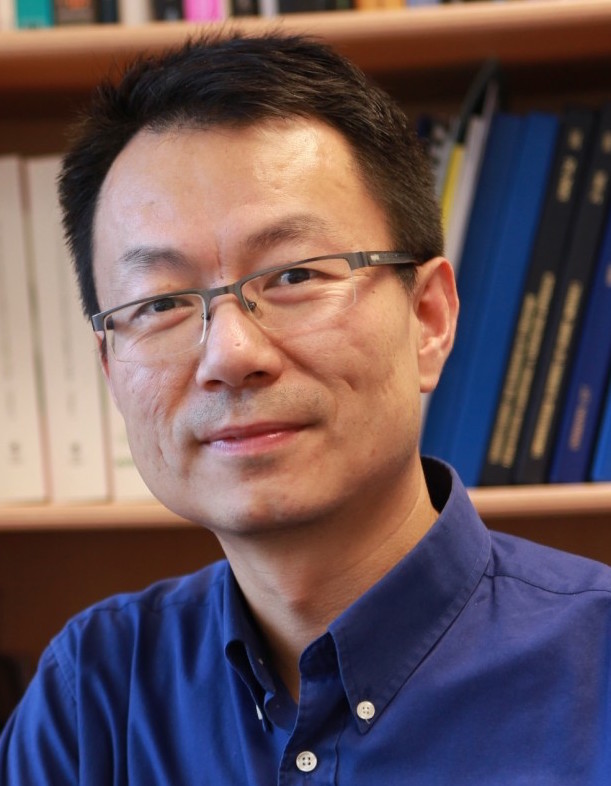}}]{Tao Xiang} received the PhD degree in electrical and computer engineering from the National University of Singapore in 2001. He is currently a Professor in the Department of Electrical and Electronic Engineering, University of Surrey and a Research Scientist at Facebook AI. His research interests include computer vision, machine learning, and data mining. He has published over 150 papers in international journals and conferences.
\end{IEEEbiography}

\begin{IEEEbiography}
[{\includegraphics[width=1in,height=1.25in,clip]{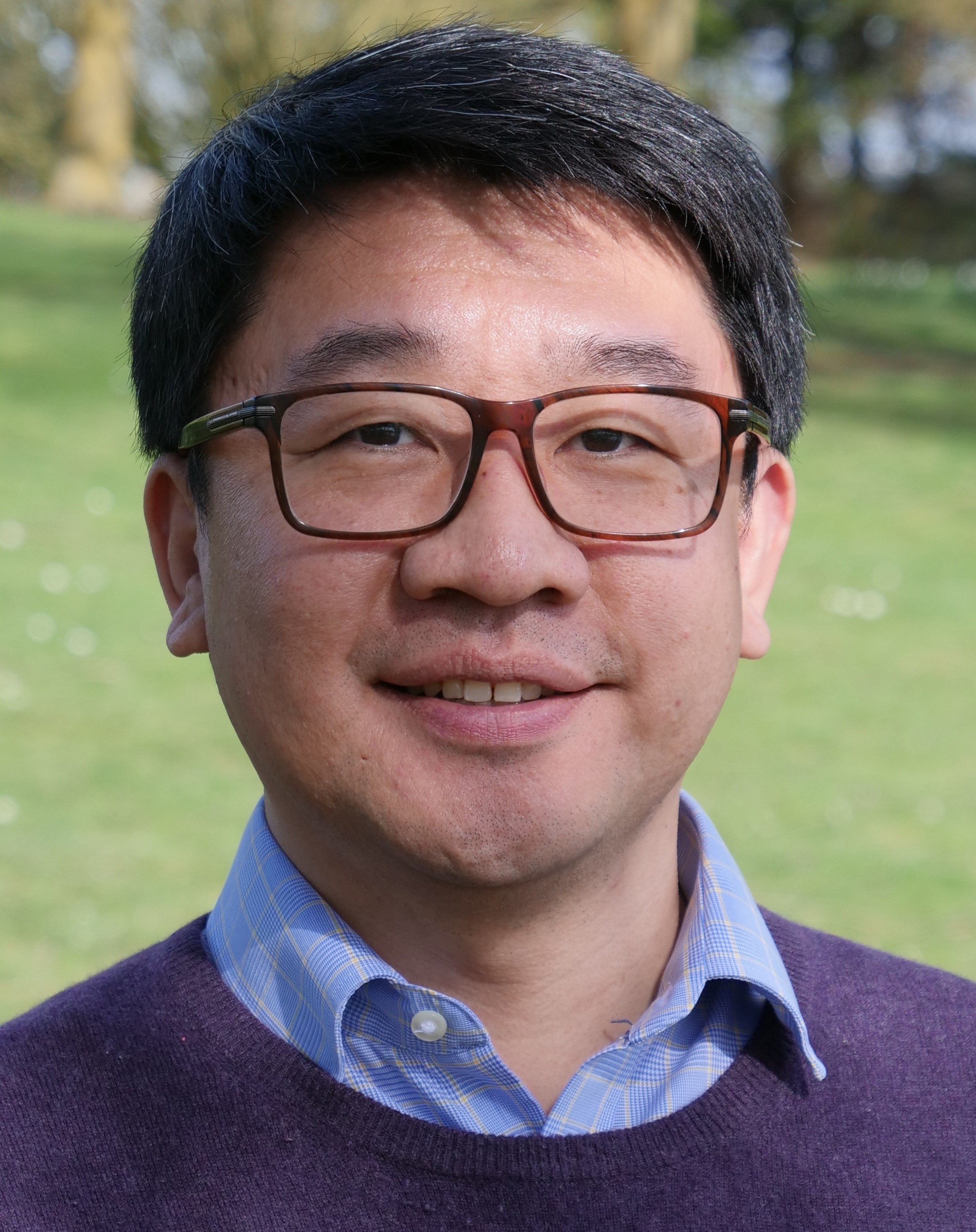}}] {Yi-Zhe Song} Yi-Zhe Song is a Professor of Computer Vision and Machine Learning, and Director of SketchX Lab at the Centre for Vision Speech and Signal Processing (CVSSP), University of Surrey. Previously, he was a Senior Lecturer at the Queen Mary University of London, and a Research and Teaching Fellow at the University of Bath. He obtained his PhD in 2008 on Computer Vision and Machine Learning from the University of Bath, and received a Best Dissertation Award from his MSc degree at the University of Cambridge in 2004, after getting a First Class Honours degree from the University of Bath in 2003. He is a Program Chair for British Machine Vision Conference (BMVC) 2021, and regularly serves as Area Chair (AC) for flagship computer vision and machine learning conferences, most recently at ICCV'21, and BMVC'20. He is a Senior Member of IEEE, a Fellow of the Higher Education Academy, as well as full member of the EPSRC review college, the UK's main agency for funding research in engineering and the physical sciences.
\end{IEEEbiography}

\clearpage
\renewcommand\appendixname{Supplemental}
\appendix

\section{Supplemental web-pages} 
We provide the supplemental web-pages that show the 5 templates, used for each category to compute the results in Table~1
in the main document, and the representative segmentation results for each method.

\section{One-shot segmentation} 
\subsection{Alternative evaluation}
In this section, we provide additional evaluation results to those in 
Section IV-D in the main document.
%
%
We provide in Table \ref{tab:main_restrictive} a more restrictive evaluation on subsets of sketches that have the same set of labels as an exemplar. Compared to the evaluation in the main paper, numerical results in Table \ref{tab:main_restrictive} do not account for the cases when the target sketch has less parts and only a part of labels has to be transferred. This is the reason why in the main document we use a less restrictive evaluation strategy.
It can be seen that similarly to the results in the main document our approach outperforms the alternative solutions.

The remaining experiments in this document use the evaluation strategy used in the main document.


\begin{table*}[t]
\centering
\footnotesize
\resizebox{\linewidth}{!}{%
\begin{tabular}{l|rr|rr|rr|rr|rr|rrrrrrrrrr}
                                & \multicolumn{10}{c|}{\textbf{P-metric}}                                                                                                                                                                                                                                                                                     & \multicolumn{10}{c}{\textbf{C-metric}}                                                                                                                                                                                                                                                                                                                                   \\ \cline{2-21} 
                                & \multicolumn{2}{c|}{\textbf{SPGG \cite{li2018universal}}} & \multicolumn{2}{c|}{\textbf{SGCN \cite{yang2020sketchgcn}}} & \multicolumn{2}{c|}{\textbf{FLSS \cite{wang2020few}}}     & \multicolumn{2}{c|}{\textbf{ISPP \cite{chen2020unsupervised}}} & \multicolumn{2}{c|}{\textbf{Ours}}                        & \multicolumn{2}{c}{\textbf{SPGG \cite{li2018universal}}}            & \multicolumn{2}{c|}{\textbf{SGCN \cite{yang2020sketchgcn}}}          & \multicolumn{2}{c|}{\textbf{FLSS \cite{wang2020few}}}                     & \multicolumn{2}{c|}{\textbf{ISPP \cite{chen2020unsupervised}}}            & \multicolumn{2}{c}{\textbf{Ours}}                        \\
\hline
\end{tabular}
}
\caption{Numerical evaluation on the SPG dataset \cite{li2018universal}: first 25 categories; on the 'airplane' category from TUBerlin \cite{eitz2012humans} and Huang14 \cite{huang2014data} datasets; on creative birds \cite{ge2020creative}. $\mu$ denotes the average accuracy over 5 runs with 5 randomly chosen templates, and $\sigma$ is the standard deviation of the 5 runs results. The evaluation in this table is done only on those sketches that have the same semantic parts as an exemplar sketch.}
\label{tab:main_restrictive}
\end{table*}

\subsection{Detailed numerical evaluation after label refinement}


In Table \ref{tab:results_refined} we provide the detailed numerical results per category. While on average our method outperforms competing approaches after refinement, our method is outperformed by ISPP method on the `bulldozer' category and tightly follows FLSS on the `suitcase' category.
The worse performance of our method than the ISPP method  on the `bulldozer' category can be explained by the fact that we solve jointly for the keypoints and stroke-level transformations. In this case, the prediction of keypoints sometimes can degrade, resulting in the method not being able to correctly estimate the global reflection between the two sketches, \eg `bulldozer' facing right or left. In Section \ref{sec:separate_train} we evaluate a separate training strategy, where the keypoints prediction network is trained separately. While separate training does increase the performance on the `bulldozer' category by $15.3$ points, in overall, the joint training strategy results in more stable performance across categories, showing better results on more categories. Please see Section \ref{sec:separate_train} for the further comparison of these two strategies.

\begin{table*}[t]
\centering
\footnotesize
\resizebox{\linewidth}{!}{%
\begin{tabular}{l|rr|rr|rr|rr|rr|rrrrrrrrrr}
                                       & \multicolumn{10}{c|}{\textbf{P-metric}}                                                                                                                                                                                                                                                                          & \multicolumn{10}{c}{\textbf{C-metric}}                                                                                                                                                                                                                                                                                                                                               \\ \cline{2-21} 
                                       & \multicolumn{2}{c|}{\textbf{SPGG \cite{li2018universal}}} & \multicolumn{2}{c|}{\textbf{SGCN \cite{yang2020sketchgcn}}} & \multicolumn{2}{c|}{\textbf{FLSS \cite{wang2020few}}}     & \multicolumn{2}{c|}{\textbf{ISPP \cite{chen2020unsupervised}}} & \multicolumn{2}{c|}{\textbf{Ours}}                        & \multicolumn{2}{c}{\textbf{SPGG \cite{li2018universal}}}                     & \multicolumn{2}{c|}{\textbf{SGCN \cite{yang2020sketchgcn}}}                  & \multicolumn{2}{c|}{\textbf{FLSS \cite{wang2020few}}}                        & \multicolumn{2}{c|}{\textbf{ISPP \cite{chen2020unsupervised}}}               & \multicolumn{2}{c}{\textbf{Ours}}                        \\
\hline
\end{tabular}
}
\caption{Numerical evaluation on the SPG dataset \cite{li2018universal}: first 25 categories; on the 'airplane' category from TUBerlin \cite{eitz2012humans} and Huang14 \cite{huang2014data} datasets; on creative birds \cite{ge2020creative}. $\mu$ denotes the average accuracy over 5 runs with 5 randomly chosen templates, and $\sigma$ is the standard deviation of the 5 runs results. The results after refining each point label with a label dominant for each stroke.}
\label{tab:results_refined}
\end{table*}

\subsection{One-shot vs. few-shot}
\label{sec:one_vs_few_sup}
In the main paper we show in Table~II that the performance improves if there are several templates available, and our results consistently outperform SGCN. 
Here in Table \ref{tab:one_vs_few_supp} we show the numerical evaluation per category.

\begin{table}[ht]
 \begin{center}
\footnotesize
\resizebox{\linewidth}{!}{%
\begin{tabular}{cl|rr|rr|rr}
\multicolumn{1}{l}{}                                                &                   & \multicolumn{2}{c|}{\textbf{1 template}}           & \multicolumn{2}{c|}{\textbf{3 templates}}          & \multicolumn{2}{c}{\textbf{5 templates}}         \\ \hline
\multicolumn{1}{l|}{}                                               & \textbf{Category} & \multicolumn{1}{c}{(P)} & \multicolumn{1}{c|}{(C)} & \multicolumn{1}{c}{(P)} & \multicolumn{1}{c|}{(C)} & \multicolumn{1}{c}{(P)} & \multicolumn{1}{c}{(C)} \\ \hline
\multicolumn{1}{c|}{\multirow{5}{*}{Ours}}                          & ambulance         & \textbf{86.0}           & \textbf{77.4}            & \textbf{92.8}           & \textbf{90.5}            & \textbf{93.2}           & \textbf{91.1}           \\
\multicolumn{1}{c|}{}                                               & apple             & \textbf{83.4}           & \textbf{69.6}            & \textbf{89.1}           & \textbf{81.3}            & \textbf{89.8}           & \textbf{80.3}           \\
\multicolumn{1}{c|}{}                                               & duck              & \textbf{88.2}           & \textbf{78.3}            & \textbf{91.3}           & \textbf{84.8}            & \textbf{92.9}           & \textbf{89.0}           \\
\multicolumn{1}{c|}{}                                               & face              & \textbf{86.0}           & \textbf{77.6}            & \textbf{91.9}           & \textbf{85.0}            & \textbf{93.8}           & \textbf{89.9}           \\
\multicolumn{1}{c|}{}                                               & pig               & \textbf{85.2}           & \textbf{76.8}            & \textbf{91.4}           & \textbf{85.6}            & \textbf{92.3}           & \textbf{88.8}           \\ \hline
\multicolumn{1}{c|}{\multirow{5}{*}{SGCN \cite{yang2020sketchgcn}}} & ambulance         & 77.5                    & 63.4                     & 90.1                    & 86.3                     & 92.1                    & 87.7                    \\
\multicolumn{1}{c|}{}                                               & apple             & 82.1                    & 65.3                     & 84.0                    & 72.0                     & 86.0                    & 77.0                    \\
\multicolumn{1}{c|}{}                                               & duck              & 74.0                    & 59.0                     & 86.2                    & 80.8                     & 77.2                    & 68.8                    \\
\multicolumn{1}{c|}{}                                               & face              & 78.4                    & 64.6                     & 86.8                    & 78.9                     & 89.9                    & 82.1                    \\
\multicolumn{1}{c|}{}                                               & pig               & 77.4                    & 64.7                     & 85.6                    & 79.2                     & 87.2                    & 80.4                   
\end{tabular}
}
 \caption{One shot vs.~few shot. See Sec.\ref{sec:one_vs_few_sup} for the details.
 }
  \label{tab:one_vs_few_supp}
    \end{center}
    \vspace{-10pt}
\end{table}

\section{Ablation studies}
\subsection{ISPP: GCN vs PointNet++ encoder}

Table {\ref{tab:ISSP_original}} shows that when the PointNet++ encoder is used as was proposed in the original paper, the ISPP method performance on one shot sketch segmentation consistently drops: The point accuracy reduces on average over the five categories by $3.6$ points, and the component accuracy -- by $5$ points.

\begin{table}[h]
\centering
\resizebox{\linewidth}{!}{%
\begin{tabular}{l|rr|rr|rr|rr}
                  & \multicolumn{4}{c|}{\textbf{P-metric}}                                                                                                                    & \multicolumn{4}{c}{\textbf{C-metric}}                                                                                                                   \\ \cline{2-9} 
                  & \multicolumn{2}{c|}{\textbf{\begin{tabular}[c]{@{}c@{}}ISPP \\ PointNet++\end{tabular}}}                               & \multicolumn{2}{c|}{\textbf{ISPP GCN}}                                       & \multicolumn{2}{c|}{\textbf{\begin{tabular}[c]{@{}c@{}}ISPP \\ PointNet++\end{tabular}}} & \multicolumn{2}{c}{\textbf{ISPP GCN}}                                     \\ \hline
\textbf{Category} & \multicolumn{1}{c}{\textbf{$\mu$}} & \multicolumn{1}{c|}{\textbf{$\sigma$}} & \multicolumn{1}{c}{\textbf{$\mu$}} & \multicolumn{1}{c|}{\textbf{$\sigma$}} & \multicolumn{1}{c}{\textbf{$\mu$}} & \multicolumn{1}{c|}{\textbf{$\sigma$}} & \multicolumn{1}{c}{\textbf{$\mu$}} & \multicolumn{1}{c}{\textbf{$\sigma$}} \\ \hline
ambulance         & 58.3                               & {\color[HTML]{9B9B9B} 8.2}             & 60.1                               & 10.3                                   & 28.5                               & {\color[HTML]{9B9B9B} 10.8}            & 33.7                               & {\color[HTML]{9B9B9B} 8.7}            \\
apple             & 73.9                               & {\color[HTML]{9B9B9B} 9.9}             & 78.2                               & 7.6                                    & 50.5                               & {\color[HTML]{9B9B9B} 18.3}            & \textbf{56.7}                      & {\color[HTML]{9B9B9B} 14.9}           \\
duck              & 67.2                               & {\color[HTML]{9B9B9B} 10.6}            & 71.2                               & 6.0                                    & 45.9                               & {\color[HTML]{9B9B9B} 14.8}            & 48.4                               & {\color[HTML]{9B9B9B} 10.5}           \\
face              & 39.2                               & {\color[HTML]{9B9B9B} 8.2}             & 41.8                               & 10.8                                   & 11.0                                 & {\color[HTML]{9B9B9B} 6.2}             & 16.6                               & {\color[HTML]{9B9B9B} 11.5}           \\
pig               & 40.7                               & {\color[HTML]{9B9B9B} 9.1}             & 45.8                               & 9.7                                    & 15.1                               & {\color[HTML]{9B9B9B} 7.5}             & 20.6                               & {\color[HTML]{9B9B9B} 8.6}            \\ \hline
Average           & 55.9                              & {\color[HTML]{9B9B9B} 9.2}             & 59.4                               & 8.9                                    & 30.2                               & {\color[HTML]{9B9B9B} 11.5}           & 35.2                               & {\color[HTML]{9B9B9B} 10.8}           \\ \hline
\end{tabular}%
}
\caption{Segmentation accuracy comparison for the ISPP \cite{chen2020unsupervised} method, when the originally proposed PointNet++ encoder is used instead of our GCN encoder.}
\label{tab:ISSP_original}
\end{table}


\subsection{Segmentation module}
As we mention in Section \ref{sec:implemntation_details} in the main document:

At inference, to obtain the labeling via Eq.~\ref{eq:segment}, we first estimate our hierarchical deformation, then the label of a point $v_i$ is obtained as follows $\tau(v_i) = \tau_{\theta_3}(v_i, \mathcal{F}^{stroke}_{s_j\in \hat{X}: v_i \in s_j}, \mathcal{F}^{sketch}_{\hat{E}})$.

Here we compare this strategy with the strategy of passing in an encoding of a globally warped target sketch $\mathcal{F}^{sketch}_{\hat{X}}$, instead of an encoding of a stroke-level warped exemplar  $\mathcal{F}^{sketch}_{\hat{E}}$. Table \ref{tab:global_encoding} shows that this strategy slightly loses the one we use in the main paper.

\begin{table}[ht]
\centering
\footnotesize
\resizebox{\linewidth}{!}{%
\begin{tabular}{l|rr|rr|rr|rr}
          & \multicolumn{4}{c|}{P-metric}                                                                                         & \multicolumn{4}{c}{C-metric}                                                                                        \\ \cline{2-9} 
          & \multicolumn{2}{l|}{$\mathcal{F}^{sketch}_{\hat{X}}$}    & \multicolumn{2}{l|}{$\mathcal{F}^{sketch}_{\hat{E}}$}    & \multicolumn{2}{l|}{$\mathcal{F}^{sketch}_{\hat{X}}$}    & \multicolumn{2}{l}{$\mathcal{F}^{sketch}_{\hat{E}}$}    \\ \hline
Category  & \multicolumn{1}{c}{$\mu$} & \multicolumn{1}{c|}{$\sigma$} & \multicolumn{1}{c}{$\mu$} & \multicolumn{1}{c|}{$\sigma$} & \multicolumn{1}{c}{$\mu$} & \multicolumn{1}{c|}{$\sigma$} & \multicolumn{1}{c}{$\mu$} & \multicolumn{1}{c}{$\sigma$} \\ \hline
ambulance & 86.7                      & {\color[HTML]{9B9B9B} 4.3}    & \textbf{87.1}             & {\color[HTML]{9B9B9B} 3.7}    & 80.2                      & {\color[HTML]{9B9B9B} 7.5}    & \textbf{81.4}             & {\color[HTML]{9B9B9B} 6.2}   \\
apple     & 94.1                      & {\color[HTML]{9B9B9B} 5.6}    & \textbf{94.3}             & {\color[HTML]{9B9B9B} 5.4}    & 86.6                      & {\color[HTML]{9B9B9B} 12.5}   & \textbf{87.1}             & {\color[HTML]{9B9B9B} 11.9}  \\
duck      & 89.1                      & {\color[HTML]{9B9B9B} 4.5}    & \textbf{89.6}             & {\color[HTML]{9B9B9B} 4.1}    & 83.6                      & {\color[HTML]{9B9B9B} 7.6}    & \textbf{84.0}             & {\color[HTML]{9B9B9B} 7.5}   \\
face      & 82.2                      & {\color[HTML]{9B9B9B} 6.4}    & \textbf{83.3}             & {\color[HTML]{9B9B9B} 6.9}    & 70.1                      & {\color[HTML]{9B9B9B} 11.5}   & \textbf{72.2}             & {\color[HTML]{9B9B9B} 11.7}  \\
pig       & 76.4                      & {\color[HTML]{9B9B9B} 16.4}   & \textbf{76.6}             & {\color[HTML]{9B9B9B} 14.6}   & 64.4                      & {\color[HTML]{9B9B9B} 20.0}   & \textbf{64.5}             & {\color[HTML]{9B9B9B} 18.1}  \\ \hline
\end{tabular}
}
\caption{Numerical evaluation of alternative strategies in the segmentation module.}
\label{tab:global_encoding}
\end{table}

\subsection{Chamfer distance in the stroke-level deformation}
Finally, we evaluate the role of the Chamfer distance in Equation~6. 
Table \ref{tab:keypoints_only}
shows the segmentation accuracy if the stroke level-deformation is guided only by the mean square distance between the keypoints of the deformed template $\hat{E}$ and the keypoints of the globally deformed sketch $\hat{X}$:  $\mathcal{L}_{MSE}(K_{\hat{E}_{\hat{X}}}, K_{\hat{X}})$.  It can be seen that using both losses $\mathcal{L}_{MSE}(K_{\hat{E}_{\hat{X}}}, K_{\hat{X}})$ and $\mathcal{L}_{CD}(\hat{X}, \hat{E}_{\hat{X}})$ gives a slight advantage over using the keypoints loss only.

\begin{table}[h]
\centering
\resizebox{\linewidth}{!}{%
\begin{tabular}{l|rr|rr|rrrr}
\textbf{}         & \multicolumn{4}{c|}{\textbf{P-metric}}                                                                                                                    & \multicolumn{4}{c}{\textbf{C-metric}}                                                                                                                    \\ \cline{2-9} 
\textbf{}         & \multicolumn{2}{c|}{\textbf{No $\mathcal{L}_{CD}$}}                         & \multicolumn{2}{c|}{\textbf{Ours full}}                                     & \multicolumn{2}{c|}{\textbf{No $\mathcal{L}_{CD}$}}                          & \multicolumn{2}{c}{\textbf{Ours full}}                                    \\ \hline
\textbf{Category} & \multicolumn{1}{c}{\textbf{$\mu$}} & \multicolumn{1}{c|}{\textbf{$\sigma$}} & \multicolumn{1}{c}{\textbf{$\mu$}} & \multicolumn{1}{c|}{\textbf{$\sigma$}} & \multicolumn{1}{c}{\textbf{$\mu$}} & \multicolumn{1}{c|}{\textbf{$\sigma$}} & \multicolumn{1}{c}{\textbf{$\mu$}} & \multicolumn{1}{c}{\textbf{$\sigma$}} \\ \hline
ambulance         & \textbf{88.4}                      & 2.4                                    & 87.1                               & 3.7                                    & \textbf{82.5}                      & \multicolumn{1}{r|}{4.3}               & 81.4                               & 6.2                                   \\
apple             & 90.8                               & 7.0                                    & \textbf{94.3}                      & 5.4                                    & 79.9                               & \multicolumn{1}{r|}{14.2}              & \textbf{87.1}                      & 11.9                                  \\
duck              & 83.1                               & 11.2                                   & \textbf{89.6}                      & 4.1                                    & 74.7                               & \multicolumn{1}{r|}{15.5}              & \textbf{84.0}                      & 7.5                                   \\
face              & 81.5                               & 6.9                                    & \textbf{83.3}                      & 6.9                                    & 69.3                               & \multicolumn{1}{r|}{8.2}               & \textbf{72.2}                      & 11.7                                  \\
pig               & 75.9                               & 15.8                                   & \textbf{76.6}                      & 14.6                                   & 62.4                               & \multicolumn{1}{r|}{20.0}              & \textbf{64.5}                      & 18.1                                  \\ \hline
Average           & 83.9                               & 8.7                                    & \textbf{86.2}                      & 6.9                                    & 73.8                               & \multicolumn{1}{r|}{12.4}              & \textbf{77.8}                      & 11.1                                  \\ \hline
\end{tabular}%
}
\caption{The role of Chamfer distance for stroke-level deformation estimation.}
\label{tab:keypoints_only}
\end{table}

\subsection{Two steps training: Isolated training for keypoints }
\label{sec:separate_train}
In this section we evaluate the overall performance of our method, if we train in two steps. 
First, we train a keypoints estimation module with our GCN sketch encoder. 
Then, we train the deformation and segmentation modules. 
In this case the GCN encoders are trained separately at each step. 
Table \ref{tab:joint_separate} provides the comparison between SGCN \cite{yang2020sketchgcn}, FLSS \cite{wang2020few}, ISPP \cite{chen2020unsupervised}, ours joint training strategy used in the main document (Ours Joint), and a two steps training (Ours Separate). 
It can be seen that on average separate training results in a slightly better average segmentation accuracy with P-metric of $84\%$  vs. $83.9\%$, and C-metric of $77.6\%$ vs. $77.4\%$. Nevertheless, (Ours Joint) strategy gives higher points accuracy than (Ours Separate) on 14 out of 25 categories on the SPG dataset. Moreover, (Ours Joint) consistently outperforms all other methods, while (Ours Separate) gives lower accuracy than SGCN on the 'backpack' and 'house' categories. 
We observe that the stroke-level deformation benefits from joint training, although, for some categories, it comes at cost of decreased performance of the keypoints prediction step (\eg the `bulldozer' category). 
Joint strategy results in a more robust performance across the categories with the standard deviation of point accuracy equal to $9.6\%$ versus $10.1\%$ for the separate training strategy (Table \ref{tab:joint_separate}).

\begin{table*}[]
\centering
\resizebox{\textwidth}{!}{%
\begin{tabular}{l|rr|rr|rr|rr|rr|rrrrrrrrrr}
                  & \multicolumn{10}{c|}{\textbf{P-metric}}                                                                                                                                                                                                                                                                                                                                   & \multicolumn{10}{c}{\textbf{C-metric}}                                                                                                                                                                                                                                                                                                                                    \\ \cline{2-21} 
                  & \multicolumn{2}{c}{\textbf{SGCN \cite{yang2020sketchgcn}}} & \multicolumn{2}{c|}{\textbf{FLSS \cite{wang2020few}}}     & \multicolumn{2}{c|}{\textbf{ISPP \cite{chen2020unsupervised}}} & \multicolumn{2}{c|}{\textbf{\begin{tabular}[c]{@{}c@{}}Ours \\ (Joint)\end{tabular}}} & \multicolumn{2}{c|}{\textbf{\begin{tabular}[c]{@{}c@{}}Ours \\ (Separate)\end{tabular}}} & \multicolumn{2}{c}{\textbf{SGCN \cite{yang2020sketchgcn}}} & \multicolumn{2}{c|}{\textbf{FLSS \cite{wang2020few}}}     & \multicolumn{2}{c|}{\textbf{ISPP \cite{chen2020unsupervised}}} & \multicolumn{2}{c|}{\textbf{\begin{tabular}[c]{@{}c@{}}Ours \\ (Joint)\end{tabular}}} & \multicolumn{2}{c}{\textbf{\begin{tabular}[c]{@{}c@{}}Ours \\ (Separate)\end{tabular}}} \\ \hline
\textbf{Category} & \multicolumn{1}{l}{$\mu$}  & \multicolumn{1}{c|}{$\sigma$} & \multicolumn{1}{l}{$\mu$} & \multicolumn{1}{c|}{$\sigma$} & \multicolumn{1}{l}{$\mu$}    & \multicolumn{1}{c|}{$\sigma$}   & \multicolumn{1}{l}{$\mu$}               & \multicolumn{1}{c|}{$\sigma$}              & \multicolumn{1}{l}{$\mu$}                 & \multicolumn{1}{c|}{$\sigma$}                & \multicolumn{1}{c}{$\mu$}  & \multicolumn{1}{c|}{$\sigma$} & \multicolumn{1}{l}{$\mu$} & \multicolumn{1}{c|}{$\sigma$} & \multicolumn{1}{l}{$\mu$}    & \multicolumn{1}{c|}{$\sigma$}   & \multicolumn{1}{l}{$\mu$}               & \multicolumn{1}{c|}{$\sigma$}              & \multicolumn{1}{l}{$\mu$}                 & \multicolumn{1}{c}{$\sigma$}                 \\ \hline
airplane          & 66.6                       & 14.0                          & 56.3                      & 11.2                          & 50.8                         & 14.2                            & \textbf{86.0}                           & 4.9                                        & 85.3                                      & 7.7                                          & 57.3                       & \multicolumn{1}{r|}{17.8}     & 34.6                      & \multicolumn{1}{r|}{15.5}     & 23.0                         & \multicolumn{1}{r|}{12.7}       & \textbf{80.6}                           & \multicolumn{1}{r|}{8.5}                   & 79.6                                      & 11.3                                         \\
alarm clock       & 79.7                       & 9.9                           & 59.7                      & 10.4                          & 59.4                         & 11.5                            & \textbf{86.4}                           & 9.1                                        & 85.4                                      & 14.4                                         & 68.4                       & \multicolumn{1}{r|}{15.7}     & 36.5                      & \multicolumn{1}{r|}{17.4}     & 32.9                         & \multicolumn{1}{r|}{17.8}       & 76.0                                    & \multicolumn{1}{r|}{15.7}                  & \textbf{77.1}                             & 22.3                                         \\
ambulance         & 78.1                       & 3.4                           & 61.5                      & 12.0                          & 60.1                         & 10.3                            & 87.1                                    & 3.7                                        & \textbf{89.1}                             & 4.6                                          & 66.9                       & \multicolumn{1}{r|}{7.2}      & 46.2                      & \multicolumn{1}{r|}{10.4}     & 33.7                         & \multicolumn{1}{r|}{8.7}        & 81.4                                    & \multicolumn{1}{r|}{6.2}                   & \textbf{85.6}                             & 7.3                                          \\
angel             & 54.2                       & 12.0                          & 47.6                      & 9.4                           & 57.8                         & 6.0                             & \textbf{70.7}                           & 11.5                                       & 69.8                                      & 13.0                                         & 45.7                       & \multicolumn{1}{r|}{13.2}     & 22.1                      & \multicolumn{1}{r|}{10.3}     & 31.4                         & \multicolumn{1}{r|}{7.2}        & \textbf{64.9}                           & \multicolumn{1}{r|}{7.8}                   & 62.5                                      & 6.8                                          \\
ant               & 44.2                       & 17.5                          & 41.7                      & 14.5                          & 47.3                         & 12.6                            & 60.8                                    & 18.0                                       & \textbf{70.1}                             & 10.1                                         & 35.2                       & \multicolumn{1}{r|}{9.5}      & 22.5                      & \multicolumn{1}{r|}{12.6}     & 27.2                         & \multicolumn{1}{r|}{15.0}       & 50.6                                    & \multicolumn{1}{r|}{18.6}                  & \textbf{55.5}                             & 11.5                                         \\
apple             & 83.4                       & 10.7                          & 82.0                      & 8.6                           & 78.2                         & 7.6                             & \textbf{94.3}                           & 5.4                                        & 94.2                                      & 5.3                                          & 78.3                       & \multicolumn{1}{r|}{16.9}     & 59.5                      & \multicolumn{1}{r|}{13.1}     & 56.7                         & \multicolumn{1}{r|}{14.9}       & \textbf{87.1}                           & \multicolumn{1}{r|}{11.9}                  & 86.8                                      & 11.7                                         \\
backpack          & 59.2                       & 3.9                           & 35.9                      & 3.8                           & 33.7                         & 6.0                             & \textbf{64.6}                           & 9.2                                        & 55.8                                      & 13.8                                         & 46.6                       & \multicolumn{1}{r|}{2.1}      & 6.4                       & \multicolumn{1}{r|}{3.8}      & 8.0                          & \multicolumn{1}{r|}{4.5}        & \textbf{50.2}                           & \multicolumn{1}{r|}{11.0}                  & 41                                        & 13.8                                         \\
basket            & 68.7                       & 15.7                          & 65.9                      & 14.2                          & 55.2                         & 15.1                            & \textbf{79.1}                           & 10.3                                       & 72.4                                      & 14.1                                         & 61.1                       & \multicolumn{1}{r|}{12.7}     & 41.6                      & \multicolumn{1}{r|}{20.4}     & 28.8                         & \multicolumn{1}{r|}{18.2}       & \textbf{70.1}                           & \multicolumn{1}{r|}{11.3}                  & 61.5                                      & 15.5                                         \\
bulldozer         & 53.4                       & 15.7                          & 56.0                      & 9.2                           & 67.9                         & 5.1                             & 69.1                                    & 11.0                                       & \textbf{84.4}                             & 5.0                                          & 43.1                       & \multicolumn{1}{r|}{18.4}     & 38.8                      & \multicolumn{1}{r|}{11.2}     & 49.1                         & \multicolumn{1}{r|}{8.9}        & 58.5                                    & \multicolumn{1}{r|}{13.4}                  & \textbf{77.4}                             & 6.4                                          \\
butterfly         & 78.2                       & 9.3                           & 70.2                      & 7.5                           & 65.0                         & 8.1                             & \textbf{91.7}                           & 3.7                                        & 86.9                                      & 7.3                                          & 67.4                       & \multicolumn{1}{r|}{13.2}     & 54.1                      & \multicolumn{1}{r|}{9.1}      & 38.9                         & \multicolumn{1}{r|}{14.1}       & \textbf{86.2}                           & \multicolumn{1}{r|}{5.8}                   & 79.3                                      & 10.5                                         \\
cactus            & 84.6                       & 4.6                           & 41.9                      & 9.1                           & 47.7                         & 11.7                            & 89.2                                    & 6.8                                        & \textbf{91.0}                             & 5.2                                          & 80.4                       & \multicolumn{1}{r|}{8.2}      & 18.9                      & \multicolumn{1}{r|}{12.2}     & 14.1                         & \multicolumn{1}{r|}{5.6}        & 83.3                                    & \multicolumn{1}{r|}{9.4}                   & \textbf{85.7}                             & 9.5                                          \\
calculator        & 89.2                       & 4.6                           & 67.3                      & 4.1                           & 52.7                         & 12.1                            & \textbf{92.6}                           & 2.8                                        & 92.3                                      & 2.9                                          & 87.7                       & \multicolumn{1}{r|}{4.8}      & 44.2                      & \multicolumn{1}{r|}{6.2}      & 24.5                         & \multicolumn{1}{r|}{7.7}        & \textbf{90.1}                           & \multicolumn{1}{r|}{5.0}                   & 88                                        & 6.5                                          \\
campfire          & 91.2                       & 3.2                           & 80.7                      & 4.2                           & 73.5                         & 5.0                             & 93.9                                    & 1.6                                        & \textbf{95.0}                             & 1.4                                          & 88.4                       & \multicolumn{1}{r|}{4.6}      & 71.4                      & \multicolumn{1}{r|}{11.4}     & 57.1                         & \multicolumn{1}{r|}{7.2}        & 89.0                                    & \multicolumn{1}{r|}{4.2}                   & \textbf{92.8}                             & 2.8                                          \\
candle            & 89.8                       & 5.7                           & 86.7                      & 4.5                           & 85.2                         & 1.7                             & 96.3                                    & 1.7                                        & \textbf{96.7}                             & 1.5                                          & 81.0                       & \multicolumn{1}{r|}{10.0}     & 71.9                      & \multicolumn{1}{r|}{10.3}     & 67.8                         & \multicolumn{1}{r|}{7.4}        & 93.9                                    & \multicolumn{1}{r|}{2.2}                   & \textbf{94.8}                             & 2.1                                          \\
coffee cup        & 73.6                       & 10.6                          & 73.7                      & 5.9                           & 66.2                         & 7.1                             & \textbf{82.6}                           & 6.9                                        & 82.3                                      & 18.9                                         & 76.2                       & \multicolumn{1}{r|}{13.6}     & 54.6                      & \multicolumn{1}{r|}{11.8}     & 38.3                         & \multicolumn{1}{r|}{15.5}       & \textbf{81.3}                           & \multicolumn{1}{r|}{5.7}                   & 78.8                                      & 22.9                                         \\
crab              & 56.2                       & 13.8                          & 49.5                      & 10.3                          & 48.6                         & 13.2                            & \textbf{75.4}                           & 12.9                                       & 72.1                                      & 13.4                                         & 51.9                       & \multicolumn{1}{r|}{11.8}     & 27.0                      & \multicolumn{1}{r|}{8.4}      & 21.6                         & \multicolumn{1}{r|}{10.5}       & \textbf{69.9}                           & \multicolumn{1}{r|}{14.8}                  & 66.4                                      & 14.9                                         \\
drill             & 71.3                       & 8.2                           & 80.6                      & 1.9                           & 84.1                         & 1.5                             & 88.7                                    & 8.0                                        & \textbf{96.7}                             & 1.0                                          & 56.4                       & \multicolumn{1}{r|}{5.3}      & 55.1                      & \multicolumn{1}{r|}{1.9}      & 68.1                         & \multicolumn{1}{r|}{8.6}        & 79.8                                    & \multicolumn{1}{r|}{10.3}                  & \textbf{95.6}                             & 1.3                                          \\
duck              & 61.2                       & 10.5                          & 53.6                      & 4.5                           & 71.2                         & 6.0                             & 89.6                                    & 4.1                                        & \textbf{91.4}                             & 3.0                                          & 53.9                       & \multicolumn{1}{r|}{9.3}      & 26.5                      & \multicolumn{1}{r|}{8.4}      & 48.4                         & \multicolumn{1}{r|}{10.5}       & 84.0                                    & \multicolumn{1}{r|}{7.5}                   & \textbf{86.6}                             & 5.8                                          \\
face              & 69.9                       & 14.4                          & 38.3                      & 6.8                           & 41.8                         & 10.8                            & \textbf{83.3}                           & 6.9                                        & 83.0                                      & 5.0                                          & 55.2                       & \multicolumn{1}{r|}{16.0}     & 12.4                      & \multicolumn{1}{r|}{8.2}      & 16.6                         & \multicolumn{1}{r|}{11.5}       & 72.2                                    & \multicolumn{1}{r|}{11.7}                  & \textbf{72.7}                             & 6.6                                          \\
flower            & 75.6                       & 14.2                          & 62.6                      & 3.5                           & 58.1                         & 3.4                             & 88.3                                    & 2.0                                        & \textbf{89.0}                             & 4.2                                          & 72.7                       & \multicolumn{1}{r|}{11.2}     & 36.4                      & \multicolumn{1}{r|}{6.5}      & 27.3                         & \multicolumn{1}{r|}{3.7}        & 87.1                                    & \multicolumn{1}{r|}{3.2}                   & \textbf{88.4}                             & 3.4                                          \\
house             & 82.2                       & 9.3                           & 57.8                      & 10.7                          & 58.4                         & 9.3                             & \textbf{89.4}                           & 2.4                                        & 75.2                                      & 17.4                                         & 81.7                       & \multicolumn{1}{r|}{7.9}      & 34.7                      & \multicolumn{1}{r|}{13.4}     & 32.8                         & \multicolumn{1}{r|}{13.3}       & \textbf{85.3}                           & \multicolumn{1}{r|}{2.4}                   & 68                                        & 17.5                                         \\
ice cream         & 82.5                       & 5.7                           & 75.2                      & 4.3                           & 72.9                         & 1.0                             & \textbf{86.5}                           & 8.3                                        & 84.5                                      & 10.1                                         & 79.2                       & \multicolumn{1}{r|}{5.9}      & 62.2                      & \multicolumn{1}{r|}{5.6}      & 60.0                         & \multicolumn{1}{r|}{2.8}        & \textbf{80.6}                           & \multicolumn{1}{r|}{8.3}                   & 79.1                                      & 10.2                                         \\
pig               & 66.8                       & 20.8                          & 37.1                      & 12.2                          & 45.8                         & 9.7                             & 76.6                                    & 14.6                                       & \textbf{78.9}                             & 10.2                                         & 55.6                       & \multicolumn{1}{r|}{20.9}     & 14.8                      & \multicolumn{1}{r|}{10.2}     & 20.6                         & \multicolumn{1}{r|}{8.6}        & \textbf{64.5}                           & \multicolumn{1}{r|}{18.1}                  & 64.3                                      & 16.8                                         \\
pineapple         & 76.9                       & 13.5                          & 66.6                      & 5.1                           & 56.5                         & 8.4                             & 80.8                                    & 8.3                                        & \textbf{85.4}                             & 5.8                                          & 74.4                       & \multicolumn{1}{r|}{12.3}     & 43.5                      & \multicolumn{1}{r|}{11.4}     & 35.0                         & \multicolumn{1}{r|}{7.3}        & 75.8                                    & \multicolumn{1}{r|}{6.8}                   & \textbf{78.3}                             & 6.3                                          \\
suitcase          & 89.2                       & 1.6                           & 82.4                      & 5.7                           & 81.7                         & 2.6                             & \textbf{93.8}                           & 1.6                                        & 93.7                                      & 1.0                                          & 90.7                       & \multicolumn{1}{r|}{1.5}      & 72.9                      & \multicolumn{1}{r|}{7.2}      & 61.0                         & \multicolumn{1}{r|}{5.3}        & \textbf{93.2}                           & \multicolumn{1}{r|}{1.7}                   & 93.1                                      & 1.1                                          \\ \hline
\textbf{Average}  & 73.0                       & 10.1                          & 61.2                      & 7.7                           & 60.8                         & 8.0                             & 83.9                                    & 7.0                                        & \textbf{84.0}                             & 7.9                                          & 66.2                       & \multicolumn{1}{r|}{10.8}     & 40.4                      & \multicolumn{1}{r|}{10.3}     & 36.9                         & \multicolumn{1}{r|}{9.9}        & \textbf{77.4}                           & \multicolumn{1}{r|}{8.9}                   & \textbf{77.6}                             & 9.8                                          \\ \hline
Min               & 44.2                       & \multicolumn{1}{l|}{}         & 35.9                      & \multicolumn{1}{l|}{}         & 33.7                         & \multicolumn{1}{l|}{}           & \textbf{60.8}                           & \multicolumn{1}{l|}{}                      & 55.8                                      & \multicolumn{1}{l|}{}                        & 35.2                       & \multicolumn{1}{l|}{}         & 6.4                       & \multicolumn{1}{l|}{}         & 8.0                          & \multicolumn{1}{l|}{}           & \textbf{50.2}                           & \multicolumn{1}{l|}{}                      & 41                                        & \multicolumn{1}{l}{}                         \\ \hline
Max               & 91.2                       & \multicolumn{1}{l|}{}         & 86.7                      & \multicolumn{1}{l|}{}         & 85.2                         & \multicolumn{1}{l|}{}           & 96.3                                    & \multicolumn{1}{l|}{}                      & \textbf{96.7}                             & \multicolumn{1}{l|}{}                        & 90.7                       & \multicolumn{1}{l|}{}         & 72.9                      & \multicolumn{1}{l|}{}         & 68.1                         & \multicolumn{1}{l|}{}           & 93.9                                    & \multicolumn{1}{l|}{}                      & \textbf{95.6}                             & \multicolumn{1}{l}{}                         \\ \hline
Std.              & 12.9                       & \multicolumn{1}{l|}{}         & 15.5                      & \multicolumn{1}{l|}{}         & 13.6                         & \multicolumn{1}{l|}{}           & 9.6                                     & \multicolumn{1}{l|}{}                      & 10.1                                      & \multicolumn{1}{l|}{}                        & 15.6                       & \multicolumn{1}{l|}{}         & 19.0                      & \multicolumn{1}{l|}{}         & 17.2                         & \multicolumn{1}{l|}{}           & 12.2                                    & \multicolumn{1}{l|}{}                      & 13.5                                      & \multicolumn{1}{l}{}                         \\ \hline
\end{tabular}%
}
\caption{The comparison of training strategies for our proposed method. Ours (Joint) refers to the joint training strategy used in the main document. Ours (Separate) refers to a two a two-steps training strategy, where we first train the keypoints prediction network, as described in Section \ref{sec:separate_train}. We also compute the minimum average accuracy across categories (Min), the maximum average accuracy across categories (Max), and the standard deviations across categories (Std.). These numbers allow to evaluate how consistent are the segmentation results of each method across different categories.}
\label{tab:joint_separate}
\end{table*}

\subsection{Keypoints sensitivity to rotations and robustness of their prediction}
As demonstrated in the supplemental web-pages and in Fig.~\ref{fig:keypoints}, keypoints prediction is robust to rotations, not affecting the segmentation performance. The mean $\mu$ and standard deviation $\sigma$ of mean $L_2$-distances between the keypoints from the original sketch and its reflected version (after reflecting back), on the ablation categories is {$\mu=0.058$, $\sigma=0.008$}.  All sketches are normalized to fit the [-0.5,0.5] bounding box.

\begin{figure}[t]
\vspace{-8pt}
\begin{center}
   \includegraphics[width=\columnwidth]{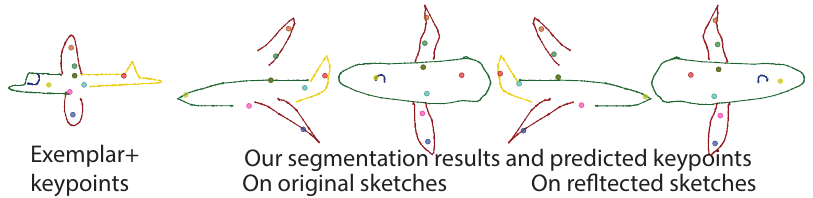}
\end{center}
\vspace{-10pt}
\caption{Keypoints and segmentation results. We visualize 8 keypoints, while use 256 for deformations computations.}
\vspace{-2pt}
\label{fig:keypoints}
\vspace{-14pt}
\end{figure}

\end{document}